\documentclass[12pt]{article}
\usepackage[polutonikogreek,latin,english]{babel}
\usepackage{tipa}
\usepackage{hyperref}
\usepackage{amsmath}
\usepackage{authblk}
\usepackage[T1]{fontenc}
\usepackage{enumitem}
\usepackage[table]{xcolor}
\usepackage{fullpage}
\usepackage{covington}
\usepackage{tikz}
\usetikzlibrary{bayesnet}
\usepackage{longtable}

\usepackage{adjustbox}
\usepackage{catchfilebetweentags}
\usepackage{microtype}
\title{Dialectal layers in West Iranian: a Hierarchical Dirichlet Process Approach to Linguistic Relationships}
\author{}
\date{}
\usepackage{setspace}
\def\citeapos#1{\citeauthor{#1}'s (\citeyear{#1})}
\usepackage{natbib}
\Roman{section}
\bibpunct[:]{(}{)}{,}{a}{}{,}
\usepackage[utf8]{inputenc}
\usepackage{pgfplots}

\newcommand{\e}[1] {\textbf{#1}}

\title{Dialectal layers in West Iranian: a Hierarchical Dirichlet Process Approach to Linguistic Relationships\footnote{To appear in {\it Transactions of the Philological Society}. Code can be found at \url{https://github.com/chundrac/w_ir_layers}.}}
\author{Chundra A.\ Cathcart\\
Department of Comparative Language Science/\\
Center for the Interdisciplinary Study of Language Evolution\\
University of Zurich\\
{\tt chundra.cathcart@uzh.ch}
}

\begin{document}

\maketitle

\begin{abstract}
\noindent 
This paper addresses a series of complex and unresolved issues in the historical phonology of West Iranian languages, (Persian, Kurdish, Balochi, and other languages), which display a high degree of irregular, non-Lautgesetzlich behavior. Most of this irregularity is undoubtedly due to language contact; I argue, however, that an oversimplified view of the processes at work has prevailed in the literature on West Iranian dialectology, with specialists assuming that deviations from an expected outcome in a given non-Persian language are due to lexical borrowing from some chronological stage of Persian. It is demonstrated that this qualitative approach yields at times problematic conclusions stemming from the lack of explicit probabilistic inferences regarding the distribution of the data: Persian may not be the sole donor language; additionally, borrowing at the lexical level is not always the mechanism that introduces irregularity. In many cases, the possibility that West Iranian languages show different reflexes in different conditioning environments remains under-explored. I employ a novel Bayesian approach designed to overcome these problems and tease apart the different determinants of irregularity in patterns of West Iranian sound change. This methodology helps to provisionally resolve a number of outstanding questions in the literature on West Iranian dialectology concerning the dialectal affiliation of certain sound changes. I outline future directions for work of this sort.
\end{abstract}

\section{Introduction}
Isoglosses based on sound changes differentiating the West Iranian languages, a group comprising Persian, Kurdish, Balochi, and other speech varieties, have long been of interest to linguists. 
The West Iranian languages are traditionally divided into Northwest (containing Kurdish, Balochi, etc.) and Southwest (containing Persian and closely related dialects) subgroups, the latter of which can be defined by a small number of phonological and morphological innovations that have taken place before the attestation of Old Persian, its oldest member. 
At the same time, a comparable (if not larger) number of Persian innovations have taken place after Old and Middle Persian, and similar innovations can be seen in other West Iranian languages, showing the effect of complex areal networks that have existed during the development of these languages. 

A number of reflexes can be identified as Southwest Iranian or Northwest Iranian on the basis of the languages in which they occur; however, language contact has complicated the picture significantly. In some cases, it is not clear what the ``correct'' outcome of a given Proto-Iranian sound should be; for instance, in the word for `spleen' (Proto-Iranian {\it *sp\textsubring{r}\'{\j}an-}), Kurdish {\it sipi\l} shows what is thought to be a typically SWIr outcome (\textit{*r/\textsubring{r}\'{\j}} $>$ {\it l}), while Persian {\it supurz} shows a typically NWIr outcome  (\textit{*r/\textsubring{r}\'{\j}} $>$ {\it rz}). 

Researchers in West Iranian dialectology have developed a set of diagnostics for marking individual words as loans in specific languages, but some of these diagnostics are better founded than others; 
in general, the picture is often so noisy, and these heuristics are so tightly intertwined, that 
all the facts cannot be qualitatively resolved within a traditional comparative-historical framework. 
I propose an alternative way of analyzing West Iranian data that integrates insights from the comparative method 
with probabilistic modeling. 
While previous research has tended to make hard decisions regarding a language's regular reflexes of sound change, this study avoids this approach; instead, I employ a quantitative approach intended to let regular behavior fall out of the data.

This paper investigates this variation in historical phonology across etymological reflexes and languages on a large scale. 
Specifically, I use a Bayesian probabilistic model, the Hierarchical Dirichlet Process (HDP), to reduce the dimensionality of the data seen within and across languages into a 
set of latent, unobserved components representing dialect membership which can be shared by multiple languages. 
The HDP is non-parametric, meaning that there is no upper bound on the number of latent features inferred. 
Both languages and phonological variants are associated with the presence of a latent feature.
This allows us to identify potential networks of language contact across the dataset.

This methodology sheds light on a number of unresolved issues in the literature on West Iranian dialectology. 
I find, unsurprisingly, that West Iranian languages show admixture (to differing degrees) from two major dialect components, roughly corresponding to Northwest and Southwest Iranian dialect groups. 
I provisionally resolve a small number of questions regarding the dialectal provenance of certain types of sound change; while the impact of this paper's results is somewhat limited due to the relatively small size of the data used, the results are interpretable, and the methodology I use is promising. 
I discuss future directions for models of this sort.

This paper is structured as follows: 
Section \ref{WIr} introduces the concept of West Iranian and the languages surveyed by this paper. 
Section \ref{WIrPhon} gives a condensed overview of the problems posed by West Iranian historical phonology for the traditional comparative method of historical linguistics. 
Section \ref{WIrPhonDetail} provides an in-depth (but not exhaustive) catalogue of sound changes in West Iranian languages, pinpointing a few problematic examples where traditional methodologies have led researchers to potentially unjustified conclusions regarding language contact within Iranian. 
Sections \ref{HDP}--\ref{inference} provide a conceptual overview of the HDP and related models, highlighting their relevance to the issues outlined in the previous section; 
an outline of the rationale for the model used in this paper; and a high-level description of the inference procedure employed. 
Results and discussion follow in sections \ref{results}--\ref{discussion}; technical details of the model employed can be found in the Appendix.

\section{The West Iranian languages}
\label{WIr}

The Iranian languages are traditionally divided into East and West subgroups, but the genetic status of these labels is shaky. 
Historically, \citet[1]{Bartholomae1883} divided Old Iranian into western and eastern variants, the former represented by Old Persian, and the latter by Avestan. The {\it Grundriss der iranischen Philologie}, particularly Wilhelm Geiger's contribution, provides a great deal of information on dialectology and subgrouping of contemporary Iranian languages. In this work, the chief distinction that cuts across Iranian is between ``Persian'' and ``Non-Persian'' dialects \citep[414]{Geiger1901}. East and West are used as purely geographic labels: at one point, Balochi, generally classified in later work as a West Iranian language, is referred to as East Iranian (p. 414). 
There is not full agreement regarding which Iranian languages are western and which are eastern; the languages Ormuri and Parachi are considered to be West Iranian languages by some scholars \citep{Grierson1918,Oranskij1977,Efimov1986}, but the consensus following \citet{Morgenstierne1929} places them in East Iranian. 
The problematic nature of the geographic labels was noted at an early date by \citet{Bailey1933}. 
\citet[651]{SimsWilliams1996} states that East Iranian is better understood as a {\it Sprachbund} than a genetic grouping, as there are very few nontrivial innovations shared by all languages in this group. 
\citet{Wendtland2009} finds that there are no secure shared phonological or morphological characteristics between the East Iranian languages, and argues against Northeast and Southeast subgroups (a division provisionally suggested in \citealt{Morgenstierne1926} and followed in \citealt{Oranskij1977}, \citealt{Kieffer1989} and elsewhere).
\citet{Cathcart2015,Korn2016,Korn2019} argue there are virtually no non-trivial innovations shared among West Iranian languages that could serve as diagnostics for subgrouping.

Regardless of the genetic status of West Iranian, the label is meaningful, not only from a typological standpoint (West Iranian languages are highly convergent in their morphosyntax) but in terms of many of the diachronic trends displayed by West Iranian languages.
Contact with non-Indo-European linguistic stocks, such as Turkic and Semitic, may have aided in shaping the linguistic profiles of West Iranian languages \citep{Stilo2005,Stilo2018}. 
Even if there are no good shared genetic innovations among West Iranian languages, the study of inter-dialectal West Iranian contact has the potential to shed light on the socio-historical development of Iran and surrounding regions.

The West Iranian languages are traditionally divided into Northwest and Southwest groups. 
The Southwest group, comprising Old, Middle and New Persian, as well as closely related dialects such as Bashkardi, Kumzari, Judeo-Tati, and others, is generally viewed as a genetic subgroup, defined by a small number of innovations. 
The Northwest group has fewer subgroup-defining innovations uniting it. 
The finer details of this distinction are not of particular importance to this paper, as the major goal here is for dialectal groupings to fall out of the behavior displayed by languages in this paper's sample, which are listed in Table \ref{langtable}.

\begin{table}
\scriptsize{
\begin{tabular}{|p{.2\linewidth}|p{.2\linewidth}|p{.2\linewidth}|p{.2\linewidth}|}
\hline
\hline
Language & Alternative names/subdialects & Sources & Glottocode\\
\hline
\hline
*$\dagger$Middle Persian (Manichean, MMP) & & \citealt{DurkinMeisterernst2004} & midd1352\\
\hline
*$\dagger$Middle Persian (Pahlavi, Phl) & {\it Book Pahlavi}, {\it Psalter Pahlavi} & \citealt{Mackenzie1971} & pahl1241\\
\hline
*Parthian (Manichean, Pth) & & \citealt{DurkinMeisterernst2004} & part1239\\
\hline
Awromani & {\it Pawai} & \citealt{BenedictsenChristensen1921} & gura1251\\
\hline
$\dagger$Bakhtiyari & & \citealt{AnonbyAsadi2014} & bakh1245\\
\hline
Marw Balochi & & \citealt{Elfenbein1963} & west2368\\
\hline
Rakhshani Balochi & & \citealt{Barker1969} & west2368\\
\hline
$\dagger$Bandari & & \citealt{Pelevin2010} & band1335\\
\hline
$\dagger$N.\ Bashkardi & & \citealt{Skjaervo1988} & bash1263\\
\hline
$\dagger$S.\ Bashkardi & & \citealt{Skjaervo1988} & bash1263\\
\hline
Zoroastrian Dari & Yazdi, Gabri (pejorative) & \citealt{Ivanow1940,VahmanAsatrian} & zoro1242\\
\hline
$\dagger$Feyli & & \citealt{MannHadank1909} & sout2640\\
\hline
Gazi & & \citealt{EilersSchapka1978} & gazi1243\\
\hline
Gilaki & {\it Rashti} & \citealt{Rastorguevaetal2012} & gila1241\\
\hline
Gurani & {\it Kandulai} & \citealt{BenedictsenChristensen1921,MannHadank1930} & gura1251\\
\hline
Kashani & & \citealt{Zhukovski1888} & kash1282\\
\hline
Khunsari & & \citealt{EilersSchapka1976} & khun1255\\
\hline
$\dagger$Kumzari & & \citealt{Thomas1930,WalAnonby2015} & kumz1235\\
\hline
Kurmanji Kurdish &  & \citealt{Soane1913,Thackstonnd} & nort2641\\
\hline
Sorani Kurdish & & \citealt{Blau1980} & cent1972\\
\hline
$\dagger$Larestani & & \citealt{Kamioka1979} & lari1253\\
\hline
$\dagger$Mamasani & & \citealt{MannHadank1909} & mama1269\\
\hline
Mazandarani & Tabari & \citealt{Nawata1984} & maza1291\\
\hline
Natanzi & & \citealt{Zhukovski1888} & nata1252\\
\hline
$\dagger$New Persian & & \citealt{Steingass1892} & midd1352\\
\hline
Qohrudi & Soi & \citealt{Zhukovski1888,Mann-Hadank1906} & khun1255\\
\hline
Sangesari & & \citealt{AzamiWindfuhr} & sang1315\\
\hline
Sivandi ($\dagger$ according to \citealt{Windfuhr2009} but not \citealt{Windfuhr1991}) & & \citealt{Lecoq1979} & siva1239\\
\hline
Taleshi & Talysh & \citealt{Schulze2000,Paul2011} & taly1247\\
\hline
$\dagger$Judeo-Tati &  & \citealt{Miller1892,Authier2012} & jude1256\\
\hline
S.\ Tati & {\it Challi, Eshtehardi, Takestani} & \citealt{Yarshater1969} & take1255\\
\hline
Vafsi & & \citealt{Stilo2004} & vafs1240\\
\hline
Vanishani & & \citealt{Zhukovski1888} & khun1255\\
\hline
Zazaki & Dimli & \citealt{Paul1998b} & diml1238\\
\hline
Zefrai & & \citealt{Zhukovski1888} & khun1255\\
\hline
\hline
\end{tabular}
}
\caption{West Iranian languages in the data set, along with alternative names and sub-variants (in italics), sources from which information was taken, and compatible glottocodes \citep{Glottolog}. Frequently used abbreviations for language names are provided. Pre-modern languages are indicated with an asterisk, Southwest Iranian languages with $\dagger$. Transcriptions for pre-modern forms follow the sources cited.}
\label{langtable}
\end{table}

\begin{figure}
\begin{adjustbox}{max totalsize={\linewidth}{\linewidth},center}
\input{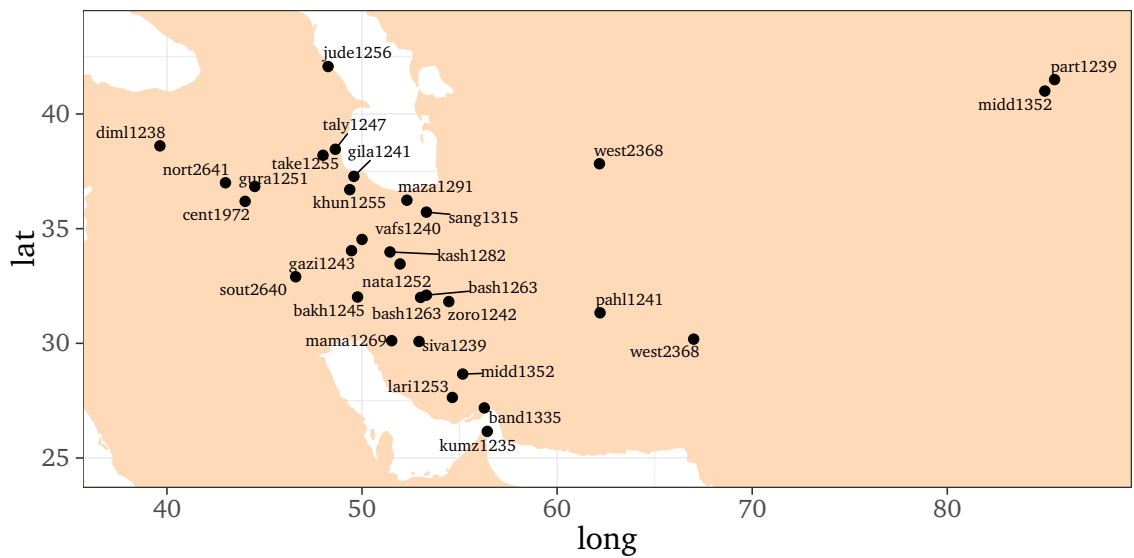}
\end{adjustbox}
\caption{Approximate locations of languages in sample}
\end{figure}

\section{West Iranian historical phonological variation}
\label{WIrPhon}

West Iranian languages show a great deal of deviation from expected outcomes of historical phonology. 
This is clear in the oldest language; Old Persian contains a number of words which display the reflexes {\it s}, {\it z}, and {\it sp} and {\it zb} for Proto-Iranian (PIr) \textit{*\'c}, \textit{*\'\j}, \textit{*\'c\textsubarch{u}}, and \textit{*\'{\j}\textsubarch{u}} instead of the expected outcomes {\it {{\texttheta}}}, {\it d}, {\it s} and {\it z}. 
This has led scholars to draw a distinction between ``proper Old Persian'' and ``Median'' forms \citep[cf.][60ff.]{Hoffmann1976}, the latter label an allusion to the confederation which preceded the Achaemenid Empire (with which Old Persian is associated). 
Although we can reliably identify only a single form as explicitly Median ({\greektext spaka} `dog', recorded by Herodotus, which shows the sound change \textit{*\'c\textsubarch{u}} > {\it sp}), a number of Old Iranian onomastic items are generally assumed to be Median. 

Words containing irregular historical phonological reflexes are common in Middle and New Persian as well, and are generally ascribed to contact with Northwest Iranian languages (although there are several probable loans from East Iranian as well; see \citealt[167]{SimsWilliams1989}). 
Northwest Iranian languages show the same degree of irregularity and contain a number of clear loans from various chronological stages of Persian, which is not surprising, given the sociopolitical influence of the Persian language in Iranian antiquity and onward. 

It is likely that a number of mechanisms have worked together to create the complex patterns seen across West Iranian. 
These include (but are not limited to) language-internal factors such as the following:
\begin{itemize}
\item Poorly understood conditioning environments: we may not fully understand the factors influencing regular sound change within languages
\item Analogical change, including paradigmatic leveling and extension, contamination, etc.
\end{itemize}
Additionally, inter-language factors like the following are almost certainly involved:
\begin{itemize}
\item Borrowing of lexical items
\item Lexical diffusion of sound changes
\end{itemize}
As mentioned above, most explanations of irregularity appeal to lexical borrowing, often from an identifiable source such as Persian. However, it is additionally possible that more than one dialectal source of similar-looking reflexes was involved (e.g., the change \textit{*\textsubarch{u}-} $>$ {\it b} may not have been restricted to Persian); furthermore, it is possible that under the umbrella of widespread multilingualism, speakers imposed sound changes from one dialect onto words from other speech varieties. In certain {\it Mischformen} it is quite clear that diffusion of sound changes was at work, rather than wholesale lexical borrowing. 
In many cases, however, it is not possible to distinguish between the two mechanisms. 
Additionally, it is not entirely clear whether similar-looking sound changes should be treated as unified, stemming from a single speech variant, or whether nearly identical sound changes were developed in parallel in different speech communities, possibly at different times. 
I identify some of these problems in the survey of sound change below, and propose a data-driven solution to at least some of these issues.

\section{West Iranian historical phonology}
\label{WIrPhonDetail}

Below, I give a synopsis of historical phonological innovations in West Iranian languages (viewed through the lens of Persian, which has the best-documented historical record) and discuss outstanding problems. 
These developments are given in rough chronological order (where a chronology can be securely established), starting with innovations preceding Old Persian, and so on, focusing on some particularly vexing problems. 

Dialectal differentiation is visible in the earliest attested West Iranian records, which consist of Achaemenid Old Persian inscriptions, as well as fragmentary Median records. 
At this stage, several phonological and morphological innovations that define Southwest Iranian as a subgroup can be identified (see below). 

The {\it locus classicus} of West Iranian dialectal differentiation is \citeapos{Tedesco1921} study of Middle Persian/Parthian isoglosses in the Manichean texts of Turfan. \citet{Lentz1926} 
discusses dialectal variation 
found in the \v{S}\=ah-n\=ama. 
In many cases this variation can be periodized with respect to when variant forms were introduced, especially in the case of Persian \citep[cf.][]{Paul2005}. 
The isoglosses identified in these works have served as the basis for a large number of dialectological investigations. Over the past century, the list of variables has been supplemented \citep{Bailey1933,Krahnke1976,Stilo1981}, and scholars have debated which features in particular are the most meaningful for West Iranian dialectology \citep{Paul1998,Korn2003,Windfuhr2009}, in terms of joint versus independent innovations.

\subsection{Changes to PIr \textit{*\'c}, \textit{*\'\j}}

The changes \textit{*\'c} $>$ {\it {{\texttheta}}} ($>$ Middle Persian, New Persian {\it h}), \textit{*\'{\j}} $>$ {\it d}, \textit{*\'c\textsubarch{u}} $>$ {\it s}, \textit{*\'{\j}\textsubarch{u}} $>$ {\it z}, \textit{*{{\texttheta}}r} $>$ {\it \c{c}},\footnote{The phonetic nature of this sound is unclear; see \citealt[287ff.]{Kuemmel2007} for a survey of possible realizations.} are found in a stratum of Old Persian (OP) vocabulary, and thought to be the expected outcome in Southwest Iranian languages. However, OP also exhibits a number of doublets or irregular reflexes of the aforementioned Proto-Iranian sounds, usually ascribed (as mentioned above) to Median admixture, though we know little about the true nature of the Median language, given the paucity of records. 

This variation is well attested in Old Persian: PIr \textit{*\'c} $>$ {\it {{\texttheta}}} in one layer of vocabulary, but {\it s} elsewhere; PIr \textit{*\'\j} $>$ {\it d} in (likely) the same stratum, but {\it z} elsewhere. This variation is described further below.

\subsubsection{PIr \textit{*\'c-}}
OP (or post-OP) initial {\it {{\texttheta}}-} consistently corresponds to Middle Persian (MP; I distinguish between Phl and MMP only for forms exhibiting variation between the two dialects) {\it s-}:\footnote{The symbol $>$, used to represent descent between forms with regular sound change, is used somewhat liberally in this paper in that it may connect OP, MP, and New Persian (NP) forms that are not actually in a relationship of direct descent (given contact); $\rightarrow$ denotes an analogical change, possibly in combination with the operation of regular sound change.}
\begin{description}
\item OP {\it a\e{\texttheta}anga-} $>$ \textit{*\e{\texttheta}anga-} : MP {\it \e{s}ang} `stone'.
\end{description}
The fact that OP {\it {{\texttheta}}} develops to MP {\it h} in most environments has led many scholars to assume that forms with MP {\it s-} are NW Iranian loans \citep[cf.][2]{Gershevitch1962}; however \citep{Salemann1901} takes initial MP {\it s-} to be the regular reflex of earlier {\it {{\texttheta}}-}. The development of PIr \textit{*\'c-} to {\it h-} is shown in a single form, NP {\it \e{h}adba} `centipede', 
found in the Burh\={a}n-i Q\={a}\d{t}\={\i}`, a 17th century dictionary 
\citep[55]{Morgenstierne1932}.

\subsubsection{PIr \textit{*\'c\textsubarch{u}}}

Reflexes of PIr \textit{*\'c\textsubarch{u}} are highly probative with respect to Iranian subgrouping; the ``proper'' Southwest Iranian outcome is taken to be {\it s}, while Khotanese Saka and Wakhi show {\it \v{s}}. 
Kurdish and Balochi, showing ``transitional'' behavior between Northwest and Southwest Iranian, appear to participate in the change \textit{*\'c\textsubarch{u}} $>$ {\it s} with Southwest Iranian, but not the changes \textit{*\'c} $>$ {\it {{\texttheta}}} $>$ {\it h}, \textit{*\'\j} $>$ {\it d}. 
In most Southwest Iranian dialects the change \textit{*\'c\textsubarch{u}} $>$ {\it s} must postdate the change \textit{*\'c} $>$ {\it {{\texttheta}}}. 
Northwest Iranian languages (other than Kurdish and Balochi) and East Iranian languages (other than Khotanese Saka and Wakhi) show 
{\it sp}, or a sequence of sounds thought to descend from it, e.g., Ossetic {\it fs}; Khunsari, Gazi, Sangesari {\it asm} `horse' $<$ \textit{*aspa-}. 
Zoroastrian Dari {\it sv} is most likely secondary rather than an archaism preserving the glide \textit{*\textsubarch{u}}, as PIr \textit{*sp} surfaces as {\it sv} as well, e.g., {\it \e{sv}arz} `spleen' $<$ \textit{*\e{sp}\textsubring{r}\'{\j}an-} \citep[21]{VahmanAsatrian}

The change \textit{*\'c\textsubarch{u}} $>$ {\it sp} cannot be reconstructed to a hypothetical ancestor of the Central Iranian languages which share it \citep[cf.][50--51]{Skjaervo2009} 
without excluding Kurdish and Balochi from this group, but these languages cannot be placed in the Southwest Iranian group: in most Southwest Iranian dialects, the change \textit{*\'c\textsubarch{u}} $>$ {\it s} must postdate the change \textit{*\'c} (which probably had a phonetic value close to {\IPA [s]}) $>$ {\it {{\texttheta}}}, as \textit{*\'c\textsubarch{u}} became \textit{*{{\texttheta}}} only in highly marginal dialects, e.g., {\it \e{t}e\v{s}} `louse' in Judeo-Shirazi (a dialect closely related to Persian but somewhat differentiated in terms of historical phonology), if from \textit{*\e{{\texttheta}}i\v{s}a-} $<$ \textit{*\e{\'c\textsubarch{u}}i\v{s}a-} \citep{Morgenstierne1960,Borjian2020}. 
It is likely that changes to \textit{*\'c\textsubarch{u}} represent a sort of areal diffusion among (originally) non-peripheral Iranian languages, albeit an old one which has operated prior to early Median and Scythian onomastic items and is found in the archaic Avestan language as well. 
It is worth noting that similar fortition of OIA {\it \'sv} has taken place in the peripheral Indo-Aryan language Khowar as well as some Nuristani languages, though \citet{Morgenstierne1926,Morgenstierne1932} cautions against connecting these developments with the Iranian one. 

Persian shows reflexes containing {\it sp} at all chronological stages. 
New Persian also shows the cluster {\it sf}. 
Henning argues that this cluster cannot be secondary from earlier \textit{*sp}, and could instead be from a dialect in which PIr \textit{*\'c\textsubarch{u}} ``resulted directly in {\it sf}'' \citep[71, fn.\ 13]{Henning1963}. 
\citet[223]{Schwartz2006} argues against the influence of Arabic (which resulted in certain sporadic {\it p} $>$ {\it f} changes, since Arabic lacks {\it p}) in certain words with {\it sf}. 
The circumstances under which NP {\it sf} came about remain unclear. 

\subsubsection{PIr \textit{*\'cr}}
A small number of Old and Middle Persian words show OP {\it \c{c}}, MP {\it s} for PIr \textit{*\'cr}, e.g., \textit{*ni-\e{\'cr}ai-} `restore' $>$ OP {\it niya\e{\c{c}}\=aray-} caus.\ (contaminated with {\it d\=araya-}, according to \citealt[188]{Kent1953}), MP {\it ni\e{s}\=ay-} `conveying, dispatch' \citep[355]{Cheung2007}. \citet[80]{Kent1942} claims that OP \textit{*\c{c}unautiy} `hear' 3sg ($<$ \textit{*\'crunauti}) yields NP {\it \v{s}un\=udan}, but the latter form is better connected with \textit{*x\v{s}nau-} `hear' \citep[456]{Cheung2007}. Elsewhere, Middle and New Persian show {\it s(V)r} and on occasion {\it \v{s}}:
\begin{description}[noitemsep]
\item PIr \textit{*\e{\'cr}\=a\textsubarch{u}a\textsubarch{i}a-} `hear' (caus.) > MP {\it \e{sr}\=ay} `sing' $>$ NP {\it \e{sur}\=u(dan)/\e{sar}\=ay} \citep[357]{Cheung2007}
\item PIr \textit{*\e{\'cr}auni-} $>$ NP {\it \e{sur}\=un} `buttocks'
\item PIr \textit{*h\textsubarch{u}a\e{\'cr}\=u-} $>$ NP {\it xu\e{sr}\=u}, {\it xu\e{\v{s}}\=u} `mother-in-law' (possibly under influence from {\it xus\=ur} `father-in-law', as suggested by a reviewer)
\item PIr \textit{*a\e{\'cr}u-(ka-)} `teardrop' $>$ Phl {\it a\e{rs}}, MMP {\it a\e{sr}}; NP {\it a\e{rs}}, {\it a\e{\v{s}}k}
\end{description}

\subsubsection{PIr \textit{*\'cn}, \textit{*\'{\j}n}}

Initial Proto-Iranian \textit{*\'cn-} appears to surface as {\it sn-} in Iranian, though evidence is restricted to reflexes of one etymon in two languages (YAv {\it \e{sn}a{\texttheta}-}, MP {\it \e{sn}\=ah-} `strike' $<$ PIr \textit{*\e{\'cn}a{\texttheta}H-}; \citealt[349]{Cheung2007}); medial \textit{*-\'cn-} $>$ OP {\it -\v{s}n-}, e.g., \textit{*\textsubarch{u}a\e{\'c-n}a-} $>$ OP {\it va\e{\v{s}n}a-} `will, favor' (OAv {\it vasn\=a}, YAv {\it vasna} `wish' inst.sg.\ may show the effects of analogy rather than a regular outcome; see \citealt[102]{HoffmannForssman2004} as well as \citealt{Schwartz2010} for an alternative view). 

PIr \textit{*\'{\j}n} appears to have become OP {\it x\v{s}n-} ($>$ MP {\it \v{s}n-} $>$ NP {\it \v{s}(V)n-}) word-initially (e.g., PIr \textit{*\e{\'{\j}n}\=a-s(\'c)a-} $>$ OP {\it \e{x\v{s}n}\=asa-} `know' $>$ MP {\it (i)\e{\v{s}n}\=as-} $>$ NP {\it \e{\v{s}in}\=as-} `recognize') and {\it -\v{s}n-} word-medially (e.g., \textit{*\textsubarch{i}a\e{\'{\j}n}a-} $>$ Phl {\it \v{\j}a\e{\v{s}n}} $>$ NP {\it ja\e{\v{s}n}} `festival', \textit{*ga\e{\'{\j}n}a-} $>$ NP {\it ga\e{\v{s}n}} `abundance'). 
From what we can tell, Northwest Iranian languages appear to have medial {\it -zn-} (on metathesis to {\it -nz-} in Median onomastic items, see \citealt{Gershevitch1962}), e.g., Parthian {\it ga\e{zn}} `treasure' vs.\ NP {\it ga\e{\v{s}}n} `abundance' $<$ \textit{*ga\e{\'{\j}n}a-}; Persian shows some forms with {\it -zn-}, possibly loans from Northwest or East Iranian, e.g., NP {\it gavazn}, cf. Sogdian {\it $\gamma$(')wzn-}, Khotanese Saka {\it gg\=uyzna-} \citep[57]{Gershevitch1954}. 
At the same time, Northwest Iranian languages appear to agree with Persian in reflecting \textit{(x)\v{s}n-} for initial \textit{*\'{\j}n-}, e.g., \textit{*\e{\'{\j}n}\=a-(s)\'ca-} `know' $>$ Parthian {\it i\e{\v{s}}n\=as}, Qohrudi {\it e\e{\v{s}n}\=as-} etc.\ \citep[466]{Cheung2007}. 

\subsubsection{PIr \textit{*\'{\j}\textsubarch{u}}}

The sequence \textit{*\'{\j}\textsubarch{u}} is found in only a small number of Proto-Iranian etyma. Old Persian contains reflexes of only two of these forms, {\it patiya\e{zb}ayam} `proclaim' 1sg.\ impf.\ ($<$ \textit{*pati-a\e{\'{\j}\textsubarch{u}}a\textsubarch{i}am} $<$ \textit{*pati-a-\'{\j}uH-a\textsubarch{i}a-m}) and {\it hi\e{z}\=ana-} `tongue' ($<$ \textit{*hi\e{\'{\j}\textsubarch{u}}\=ana-}). 
The latter form is believed by most scholars to be ``proper'' OP \citep{Kent1953}, and the former a Median loan.\footnote{Verbal adjectives involving this form have found their way into A\'soka's Aramaic inscriptions \citep{Schwarzschild1960}.} 
The Middle Persian word for tongue is {\it u\e{zw}\=an/i\e{zw}\=an}, and cannot be taken as a direct reflex of the OP form (since {\it z} from other sources does not change to {\it zw}); Middle Persian however shows the development \textit{*\'{\j}\textsubarch{u}} $>$ {\it z} in {\it par\textbf{z}\={\i}r-} `keep away' $<$ \textit{*para-/pari-\e{\'{\j}\textsubarch{u}}ar-} \citep[475]{Cheung2007}. New Persian and related dialects tend to show {\it zVb} or {\it zVw} (e.g., NP {\it \e{zab}\=an}, {\it \e{zuw}\=an} `tongue' $<$ \textit{*hi\e{\'{\j}\textsubarch{u}}\=ana-}). 

\subsubsection{PIr \textit{*\'c\textsubarch{i}}}
\label{SY}
The development of PIr \textit{*\'c\textsubarch{i}} in Persian is not entirely clear. 
Its fate is intertwined with that of the cluster \textit{*{{\texttheta}}\textsubarch{i}} ($<$ PIIr \textit{*t\textsubarch{i}}), whose regular Old Persian reflex is thought to be {\it \v{s}iy} \citep[32]{Kent1953}; cf.\ \textit{*h\textsubarch{u}ai-pa\e{{{\texttheta}}\textsubarch{i}}a-} $>$ MP {\it xw\=e\e{\v{s}}} `self'.

Old Persian attests the cluster \textit{*\'c\textsubarch{i}} only word-medially, where it surfaces as {\it {{\texttheta}}iy} (showing characteristic OP anaptyxis between consonants and glides), e.g., {\it vi{{\texttheta}}iy\=a} `house' loc.sg., possibly via paradigmatic leveling of the stem {\it vi{{\texttheta}}-} ($<$ {\it *\textsubarch{u}i\'c-}). 
Old Persian does not directly attest this cluster word-initially; Middle Persian shows varying reflexes:
\begin{description}[noitemsep]
\item PIr \textit{*\e{\'c\textsubarch{i}}\=a\textsubarch{u}a-(ka-)} $>$ MP {\it \e{siy}\=ah} $>$ NP {\it \e{sy}\=ah} `black'
\item PIr \textit{*\e{\'c\textsubarch{i}}aina-(m\textsubring{r}ga-)} $>$ MP {\it \e{s}\=en murw} `a fabulous bird' $>$ NP {\it \e{s}\={\i}mur{\textgamma}} \citep[74]{Mackenzie1971}. Young Avestan {\it sa\=ena-} `eagle' may show a dissimilation \textit{*\'c\textsubarch{i}} $>$ \textit{*\'c} the presence of the off-glide of the diphthong {\it ai}, which may also account for Persian {\it s}, but this development was clearly not pan-Iranian, since the initial consonant of Balochi {\it \e{\v{s}}\=enak} `falcon, hawk' \citep[129]{Korn2005} cannot continue PIr \textit{*\'c-}.
\end{description}

Word-internally, Middle and New Persian reflect variation between earlier \textit{*\v{s}i\textsubarch{i}} and \textit{*{{\texttheta}}i\textsubarch{i}}. 
\begin{description}[noitemsep]
\item PIIr \textit{*mats\textsubarch{i}a-(ka-)} $\rightarrow$ \textit{*m\=a\e{\'c\textsubarch{i}}a-ka-} (\citealt[637, fn. 25]{Hoffmann1976} attributes the long vowel to V\textsubring{r}ddhi) $>$ MP {\it m\=a\e{h}\={\i}g} `fish' $>$ NP {\it m\=a\e{h}\={\i}}
\item PIIr \textit{*tus\e{\'c\textsubarch{i}}a-(ka-)} $>$ likely PIr \textit{*tu\e{\'c\textsubarch{i}}a-ka-} $>$ MP {\it tu\e{h}\={\i}g} `barren'
\item PIr \textit{*ka\e{\'c\textsubarch{i}}apa-ka-} $>$ MP {\it ka\e{\v{s}}avag} `tortoise' $>$ NP {\it ka\e{\v{s}}av}, {\it ka\e{\v{s}}af}; cf.\ Bandari {\it k\=a\e{s}apo\v{s}t} `turtle', Balochi {\it ka\e{s}\={\i}p} $\sim$ {\it ka\e{s}\={\i}b} `turtle, tortoise', Kurmanji {\it k\={\i}\e{s}al}, Sivandi {\it ka\e{l}apo\v{s}t} (showing the East Iranian change \textit{*\v{s}} $>$ {\it l}, according to \citealt{Asatrian2012}; cf.\ \citealt[167]{SimsWilliams1989}), Bakhtiyari {\it k\=a\e{s}epo\v{s}t} `turtle', Zazaki {\it ke\e{s}e}
\item PIr \textit{*\textsubarch{u}a\e{\'c\textsubarch{i}}ah-} $>$ MP {\it w\=e\e{\v{s}}} `more' $>$ NP {\it b\=e\e{\v{s}}} `more'; cf.\ Sivandi {\it v\={\i}\e{\v{s}}tar}, Balochi (Marw) {\it ge\e{\v{s}}tir} `greater, oftener'
\item PIr \textit{*ka\e{\'c\textsubarch{i}}ah-} $>$ MP {\it ke\e{h}} `small(er), young(er)' $>$ NP {\it ki\e{h}} `small'
\item PIr \textit{*ma\e{\'c\textsubarch{i}}ah-} $>$ MP {\it me\e{h}}, {\it ma\e{h}y} (MMP <mhy>) `great(er), old(er)' $>$ NP {\it mi\e{h}} `big'
\end{description}
Variation of this sort led \citet[19--22]{Gershevitch1962} to argue that $\theta$ was an optional pronunciation of {\it s} in Old Persian (see \citealt[cf.][637, fn.\ 25]{Hoffmann1976} for a disagreeing view). 
\citet[203]{Klingenschmitt2000} ascribes alternation between pre-OP \textit{*-i\textsubarch{i}a-} and \textit{*-\textsubarch{i}a-} to 
variation among suffixes, 
with pre-OP \textit{*{{\texttheta}}\textsubarch{i}} yielding {\it \v{s}iy} and pre-OP \textit{*{{\texttheta}}i\textsubarch{i}} yielding {\it {{\texttheta}}iy}. 
\citet{Cantera2009} invokes a rhythmic law proposed by Klingenschmitt ({\it ibid.}) to account for phonological irregularities in Middle Persian nouns. 
Armed with these ideas, we can 
account for some of the variation within Persian, 
if we assume the pre-forms 
\textit{*m\=a\e{{{\texttheta}}\'i\textsubarch{i}a}-ka-} and \textit{*tu\e{{{\texttheta}}\'i\textsubarch{i}a}-ka-} versus \textit{*ka\e{\'c\textsubarch{i}}\'apa-ka-} and \textit{*\textsubarch{u}\'a\e{\'c\textsubarch{i}}ah-} (the stress placement assumed here follows \citealt[30ff.]{Back1978}), 
but this still does not explain {\it h} in reflexes of \textit{*ka\e{\'c\textsubarch{i}}ah-} and \textit{*ma\e{\'c\textsubarch{i}}ah-}, which should have undergone the same development as \textit{*\textsubarch{u}\'a\e{\'c\textsubarch{i}}ah-}, as noted by Gershevitch. 

In Persian, as elsewhere in West Iranian, language contact, analogy, and prosodically conditioned change have interacted to 
bring about the complex variation seen in reflexes of PIr \textit{*\'c\textsubarch{i}} and related sounds. 
The limited knowledge we have of late Old Persian prosody can help to tease out the role of the last mechanism, but only to a certain extent. 
It is tempting to account for variation in West Iranian languages with no documented history in a similar manner, but this is purely speculative. 
An instructive example is the following thought experiment: 
\citet[284]{Korn2005} contends while discussing Balochi {\it k\=asib}/{\it kas\={\i}p} `turtle, tortoise' that ``a genuine Bal.\ word should show {\it \v{s}}.'' However, given that Balochi {\it \={\i}} reflects PIr \textit{*-i\textsubarch{i}\twoacc[\u|\={a}]-} \citep[cf.][105]{Korn2005}, a pre-Balochi form \textit{*ka\'ci\textsubarch{i}apa-} is not inconceivable on historical phonological grounds, but perhaps overly speculative since we know virtually nothing about the phonotactics, syllabification, and stress pattern of Balochi's precursor. 
However, we also do not know whether {\it \v{s}} is the Balochi reflex of \textit{*\'c\textsubarch{i}} across the board (as assumed by Korn), or only in specific environments. 
Ultimately, 
we may benefit from relaxing some of these assumptions and employing a probabilistic model that allows us to make generalizations regarding languages' diachronic behavior on the basis of intra-language and inter-language distributions of sound changes. 



\subsection{PIr \textit{*{{\texttheta}}}}

PIr \textit{*{{\texttheta}}} changes into OP {\it {{\texttheta}}} ($>$ MP, NP {\it h}) in most conditioning environments, though it may develop into MP {\it s-} word-initially, e.g., PIr \textit{*\e{{\texttheta}}axta-} (cf.\ Khwarezmian {\it \e{{\texttheta}}{\textgamma}d}) $>$ NP {\it \e{s}axt} `hard'. 
The change \textit{*{{\texttheta}}r > \c{c}} {\IPA [s]} is well established, as is \textit{*{{\texttheta}}\textsubarch{i}} $>$ {\it \v{s}y}, though numerous exceptions to these developments exist as well.

\subsubsection{PIr \textit{*{{\texttheta}}n}}

There are relatively few Proto-Iranian sources of the cluster \textit{*{{\texttheta}}n}, but these are realized as {\it \v{s}n} across the board in West Iranian, to the exclusion of the possible Median proper name in Akkadian {\it Pa-at-ni-e-\v{s}a-} = Med {\it *Pa{{\texttheta}}n\={\i}\textsubarch{i}\=e\v{s}a-} $<$ \textit{*pa{{\texttheta}}n\={\i}-ai\v{s}a-} `looking for a wife' \citep[273]{Tavernier2007}.\footnote{A possible but highly unlikely exception is the OP form {\it k\textsubring{r}nuvaka-} `stonemason' \citep[found in the Susa inscription of Darius, ][133, 145]{Schmitt2009}, if from \textit{*k\textsubring{r}t-nu-aka-} following \citet[80]{Kent1942}, though \citet[180]{Kent1953} is less certain regarding the presence of \textit{*-t-} and other scholars \citep[e.g.,][163]{Brust2018} make no mention of etymological {\it *-t-}; a pre-form \textit{*k\textsubring{r}-nu-aka-} is far more likely.}
\begin{description}[noitemsep]
\item PIr \textit{*ara\e{{\texttheta}n}i-} $>$ OP {\it ara\e{\v{s}n}i-} `cubit' $>$ MP {\it \=are\e{\v{s}n}} $>$ NP {\it \twoacc[\u|\={a}]re\e{\v{s}(n)}}
\item PIr \textit{*dm\=ana-pa\e{{\texttheta}n}i-} $>$ MP, Parthian {\it b\=ambi\e{\v{s}n}} `queen'
\item PIr \textit{*-i-\e{{\texttheta}n}a-} $>$ MP abstract noun suffix {\it -i\e{\v{s}n}} $>$ NP {\it -i\e{\v{s}}}; cf.\ Zazaki infinitive suffix {\it -i\e{\v{s}}} \citep[105]{Benveniste1935}
\end{description}
Middle Persian {\it \=aren\v{c}} $>$ NP {\it \=aranj} `elbow', a doublet with {\it \twoacc[\u|\={a}]re\v{s}(n)} `cubit', is most likely a loan from a source closely related to Sogdian (cf. {\it \=ari\e{n}\v{c}} $<$ \textit{*ara\e{{{\texttheta}}n}i-ka-}). If NP {\it \=amvasn\={\i}} `rival wife' is to be connected with \textit{*ham-pa{{\texttheta}}n\={\i}-} (\citealt[119]{Tafazzoli1974}; \citealt[134]{MonchiZadeh1990})
it perhaps shows a secondary change \textit{*\v{s}} $>$ {\it s} seen in some other words.

\subsection{PIr \textit{*\v{s}t}, \textit{*\v{z}d}}

Old, Middle and New Persian (along with other Iranian languages) show variation between {\it st} and {\it \v{s}t} for PIr \textit{*\v{s}t}; later language attests variation between {\it zd} and {\it \v{z}d} (e.g., MP {\it mizd} `reward', NP {\it mizd}, {\it mu\v{z}d}). 
There is disagreement as to whether OP {\it st} for {\it \v{s}t} is due to analogy \citep[34]{Kent1953} or a sound change defining Southwest Iranian \citep{Skjaervo1989}, and what the relationship of this behavior is to similar-looking developments in the later language. 
\citet[196ff.]{Lipp2009} states that OP {\it -st-} (found as a reflex of PIE \textit{*-\^{k}-t-}, \textit{*-\^{g}-t-}) is due to analogy, while other developments are due to a phonological change predating Middle Persian:

\begin{enumerate}[noitemsep]
\item PIE \textit{*h$_3$re\^{g}-to-} $>$ PIr \textit{*ra\e{\v{s}t}a-} $\rightarrow$ OP {\it r\=a\e{st}a-} `right' $>$ MP {\it r\=a\e{st}} $>$ NP {\it r\=a\e{st}}
\item PIr \textit{*mu\e{\v{s}}ti-} `fist' $>$ MP {\it mu\e{\v{s}}t}, {\it mu\e{st}} $>$ NP {\it mu\e{st}}
\item PIr \textit{*-i\e{\v{s}t}a-} (superlative suffix) $>$ MP {\it -i\e{st}}; e.g., Phl {\it b\=ali\e{st}}, MMP {\it b\=ari\e{st}} `highest' $<$ \textit{*bar\'{\j}i\e{\v{s}}ta-}; Phl {\it xw\=ali\e{st}}, MMP {\it xw\=ari\e{st}} `sweetest' $<$ \textit{*h\textsubarch{u}a{r\'{\j}}i\e{\v{s}t}a-} (cf.\ Iron Ossetic {\it xo{rz}}, Digor Ossetic {\it xua{rz}} `good')
\end{enumerate}


\subsection{\textit{*r} $+$ \textsc{coronal} change}

\subsubsection{Change to \textit{l}}

A number of West Iranian forms show a sound change whereby \textit{*r} $+$ {\sc coronal} sequences become {\it l}. This behavior is common in Middle and New Persian, perhaps representing a regular sound change which operated between Old and Middle Persian:
\begin{description}[noitemsep]
\item PIr \textit{*\'{\j}\e{\textsubring{r}d-}} $>$ MP {\it di\e{l}} $>$ NP {\it di\e{l}} `heart'
\item PIr \textit{*\textsubarch{u}(a)\e{rd}a-} $>$ MP {\it gu\e{l}} $>$ NP {\it gu\e{l}} `flower'
\item PIr \textit{*\'ca\e{rd}a-} $>$ OP {\it {\texttheta}ard-} $>$ Phl {\it s\=a\e{l}}, MMP {\it s\=a\e{r}} $>$ NP {\it s\={a}\e{l}} `year'
\item PIr \textit{*p(a)\e{rd}anku-} $>$ NP {\it pa\e{l}ang} `panther'; cf.\ Vedic {\it p\e{\twoacc[\'|\textsubring{r}]d}\=aku-} (with meaning `leopard' in the Paippal\=ada recension of the Atharva Veda, \citealt[59]{Zehnder1999}), Sogdian {\it pw\e{r$\boldsymbol{\delta}$}'nk} `panther, leopard'\footnote{This form is likely a Wanderwort, but seems reconstructible to Proto-Iranian; see \citealt{Lubotsky2001}.}
\item PIr \textit{*b\e{\textsubring{r}\'{\j}}ant-} $>$ MP {\it bu\e{l}and} $>$ NP {\it bu\e{l}and} `high'
\item PIr \textit{*na\e{rd}-} `lament, moan' $>$ MP {\it n\=a\e{l}-} $>$ NP {\it n\=a\e{l}(\={\i}dan)} `lament' \citep[282]{Cheung2007}
\item PIr \textit{*ba\e{r\'{\j}}\=ad(a)-}\footnote{Perhaps a de-instrumental {\it d-}stem built to PIE \textit{*b\textsuperscript{h}er\'g\textsuperscript{h}-eh$_1$-}, cf.\ Latin {\it merc\=es}, {\it merc\=edis} `wages', {\it her\=es}, {\it her\=edis} `heir' \citep[304-5]{Weiss2009}.}
 $>$ MP {\it b\=a\e{l}\=ay} $>$ NP {\it b\=a\e{l}\=a} `height'
\item PIr \textit{*ma\e{r\'{\j}-}} $>$ MP {\it m\=a{l}-} `rub, sweep' $>$ {\it m\=a\e{l}\={\i}dan} `rub, polish'
\item PIr \textit{*\e{\textsubring{r}\'{\j}}if\textsubarch{i}a-} $>$ Phl {\it \=aluh}, MMP {\it \=aluf} $>$ NP {\it \=a\e{l}uh} `eagle'
\item PIr \textit{*\'{\j}a\e{rn}u-mani-} `gold neck' $>$ NP {\it d\=a\e{l}-man} `black eagle' \citep[292, fn.\ 14]{Schwartz1970}
\item PIr \textit{*g(a)\e{\textsubring{r}n}a-ka-} (cf.\ Old Indic {\it ga\e{\d{n}}\'a-}) $>$ NP {\it ga\e{l(l)}a} `flock' \citep[292, fn.\ 14]{Schwartz1970}
\item PIr \textit{*p\e{\textsubring{r}t}u-} $\rightarrow$ \textit{*p\e{\textsubring{r}{\texttheta}}u-} $>$ MP {\it puhl} $>$ NP {\it p\twoacc[\u|\={u}]\e{l}} `bridge'
\item PIr \textit{*pa\e{r\'c}u-} $\rightarrow$ \textit{*par\e{{{\texttheta}}(a)\textsubarch{u}}-a-ka-} $>$ mp {\it pa\e{hl}\=ug} `side, rib' $>$ NP {\it pa\e{hl}\=u} `side'
\item PIr \textit{*\v{c}a{{\texttheta}}\textsubarch{u}(a)r-\'cat-} \citep[cf.][309]{Emmerick1992} $>$ PSWIr \textit{*\v{c}a{\texttheta}\textsubarch{u}\e{\textsubring{r}}\e{\texttheta}at-} $>$ MP {\it \v{c}ehe\e{l}} $>$ NP {\it \v{c}ihi\e{l}} `forty' \citep[309]{Emmerick1992}, Judeo-Tati {\it \v{c}\"u\e{l}} \citep[88]{Authier2012}, cf.\ Zazaki {\it \v{c}ew\e{res}} \citep[61]{Paul1998}
\end{description}
In some cases, this development has operated across an intervening vowel, likely unstressed:
\begin{description}[noitemsep]
\item PIr \textit{*\'ca\e{r(a)-d}\=ara-} \citep[194]{Klingenschmitt2000} $>$ Phl `Tr\"ager der Mund; Oberster' {\it s\=a\e{l}\=ar} (MMP {\it s\=a\e{r}\=ar} `leader') $>$ NP {\it s\=a\e{l}\=ar} `leader' (cf.\ NP {\it sa\e{r-d}\=ar}, perhaps a later compound)
\item PIr \textit{*pa\e{ri-d}\=ana-} $>$ NP {\it p\=a\e{l}\=an} `pack-saddle' (cf.\ Sogdian {\it py\e{r$\boldsymbol{\delta}$}nn} `saddle', cf.\ \citealt[181]{SimsWilliams1989})
\item PIr \textit{*pa\e{ri-d}ai\'{\j}a-} $>$ NP {\it p\=a\e{l}\=ez} `garden' \citep[53]{Cheung2007}
\end{description}
However, this development is not exceptionless: it does not operate in forms like 
NP {\it pa\e{d}arzah} `a wrapper in which clothes are folded up', if from \textit{*pa\e{ri-d}ar\'{\j}-aka-} (\citealt[63]{Cheung2007}, marked as a loanword perhaps due to {\it z} $<$ \textit{*\'{\j}}), which appears to have undergone a dissimilatory development \textit{r...r} $>$ \textit{$\emptyset$...r} that is not paralleled in \textit{*\'car(a)-d\=ara-}. 
It is unlikely that language-internal factors (viz., different conditioning environments) can account for the entire range of variation seen within Persian. 

The lateralization of \textit{*r} $+$ \textsc{coronal} sequences, from what we can see, post-dates Old Persian. 
However, there are a large number of exceptions to this rule within Middle and New Persian; for example, NP {\it bu\e{l}and} forms a doublet with {\it bu\e{rz}}, thought to represent a Northwest Iranian form \citep[3]{Beekes1997}. 
For some etyma, Persian lacks {\it l}, while a non-Persian reflex displays it, e.g., NP \textit{supu\e{rz}} `spleen' versus Kd \textit{sipi\e{\l}} `\textit{id.}' $<$ PIr \textit{*sp\e{\textsubring{r}\'{\j}}an-}. 
The uncertainty surrounding this behavior can be summed up by the following comment by \citet[78]{Mackenzie1961} on the outcome of PIr \textit{*rd/r\'{\j}} in Kurdish: ``I do not think it is possible to be certain which is the true Kurdish development, but whether we consider the many words with \textit{l/\l} as native or loan-words their preponderance is significant.''

Gurani contains the forms {\it zi\e{l}} `heart' ($<$ \textit{*\'{\j}\e{\textsubring{r}d}-}) and {\it wi\e{l}\={\i}} `flower' ($<$ \textit{*\textsubarch{u}\e{\textsubring{r}d}a-}, suffixation unclear), which cannot be Persian loans; in the first, the change to OP {\it d} predates lateralization of \textit{*r/\textsubring{r}d-}. In the second form, it is most probable that the change to {\it g-} in MP {\it gul} was triggered by the following \textit{*-\textsubring{r}-}, which subsequently underwent lateralization (e.g., \textit{*\textsubarch{u}\textsubring{r}da-} $>$ \textit{*gVrd-} $>$ {\it gul}, see \S\ref{W}).\footnote{If Semnani {\it val(a)} reflects full-grade \textit{*\textsubarch{u}arda-}, it could in theory be a Persian loan, since changes affecting MP {\it wa-} post-date the \textit{*rd} $>$ {\it l} change; however, there is no concrete evidence that Persian continues \textit{*\textsubarch{u}arda-}.} Whether {\it l} in these forms owes itself to Persian influence as opposed to some other source is unclear.

\subsubsection{\textit{*rn} $>$ \textit{rr}}
The change \textit{*rn} $>$ {\it rr} is attested in Middle Persian and onward, as seen in the following examples:\footnote{The {\it rr} sequence found in NP {\it xurram} `joyful, lucky' and some other forms may be secondary 
(cf.\ \citealt[106]{Horn1893}; \citealt[55]{Hubschmann1895}).}
\begin{description}[noitemsep]
\item PIr \textit{*h\textsubarch{u}a\e{rn}ah-}/\textit{*fa\e{rn}ah-} (on the reconstruction of this etymon see \citealt{Skjaervo1983}, \citealt{Lubotsky2002}) $>$ Phl {\it xwa\e{rr}ah}, MMP {\it farrah} $>$ NP {\it fa\e{rr}} `glory'
\item PIr \textit{*\textsubarch{u}a\e{rn}a-ka-} `lamb' $>$ MP {\it wa\e{rr}ag} $>$ NP {\it ba\e{rr}ah}
\item PIr \textit{*pa\e{rn}a-} $>$ MP {\it pa\e{rr}} $>$ NP {\it pa\e{r(r)}} `feather'
\item OP 
{\it k\e{\textsubring{r}n}uvak\={a}} {\it <k-r-n\textsuperscript{u}-u-v-k-a>} `stonemason' $>$ MP {\it ki\e{rr}\=og} `artisan' 
\item PIr \textit{*d(a)\e{r-n}-} $>$ MP {\it da\e{rr}-} $>$ NP {\it da\e{rr-}} `to rend, tear up'
\item PIr \textit{*da\e{rn}a-ka-} $>$ NP {\it da\e{rr}ah} `valley', cf. YAv {\it u\v{s}i.dar{\textschwa}na-} `having reddish cracks', name of a mountain (\citealt[124--5]{Horn1893}; \citealt[180]{HumbachIchaporia1998})
\item PIr \textit{*us-p\e{\textsubring{r}n}a-} $>$ MP {\it aspurr} `accomplished' \citep[228]{Klingenschmitt2000}
\end{description}
It has been suggested that the changes \textit{*rn} $>$ {\it rr} and \textit{*rn} $>$ {\it l} (see above) are interconnected, and that {\it l(l) $\sim$ r(r)} variation in reflexes of \textit{*rn} represents dialectal variation within West Iranian
\citep[292, fn.\ 14]{Schwartz1970}. 

Middle and New West Iranian languages as a whole show an overwhelming tendency toward the change \textit{*rn} $>$ {\it r(r)}. 
West Iranian words for `lamb', if reflexes of \textit{*\textsubarch{u}arna-(ka-)} ($<$ PIE \textit{*\textsubarch{u}\textsubring{r}h$_1$n-}; \citealt[225--6]{EWAia1}), 
show this behavior across the board:
\begin{description}[noitemsep]
\item MP {\it wa\e{rr}ag} $>$ NP {\it ba\e{rr}a}; Parthian {\it wa\e{rr}ag}; Awromani {\it v{\ae}\e{r}\={a}}; Balochi {\it gw\={a}\e{r}ag}; S Bashkardi {\it va\e{r}k}; Gurani {\it va\e{r}\={a}la}, {\it valala} ($<$ \textit{*\textsubarch{u}arna-la-}?); Zazaki {\it vo\e{r}ek} `lamb'
\end{description}
However, Balochi and Parthian forms show the change \textit{*rn} $>$ {\it n(n)}; Zazaki shows {\it rn} only via analogical maintenance or restoration, but otherwise {\it r} $\sim$ {\it \={r}} \citep[133--4]{Korn2005}.

\subsection{\textit{r} $\sim$ \textit{l} variation}
Proto-Indo-European \textit{*l} surfaces as {\it r} in the vast majority of Iranian languages. PIE \textit{*l} $>$ \textit{*r} is often given as a Proto-Iranian sound change in most handbooks, yet there are a number of exceptions to this development \citep{Schwartz2008}, indicating that PIE \textit{*l} has been conserved in some peripheral dialects. 
Northwestern dialects also contain morphological variants with {\it l} lost by Persian with congeners in Indic, e.g., Kashani {\it engul\={\i}}, Mazandarani {\it engel} (cf.\ Old Indic {\it a\.ng\'uli-}) against NP {\it angu\v{s}t} (cf.\ OInd {\it a\.ng\'u\d{s}\d{t}ha-}) `finger' (\citealt[]{Horn1893}; \citealt[226--8]{Krahnke1976}).\footnote{If the term for `shepherd' in languages spoken in the Caspian region, {\it g\=ale\v{s}}, comes, as suggested by \citet{Asatrian2002}, from \textit{*ga\textsubarch{u}a-rax\v{s}a-} `cow protector' (cf.\ Old Indic {\it go-rak\d{s}a-}), then the presence of {\it l} is in agreement with PIE \textit{*h$_2$leks-}, pointing to another possible archaism.}

However, some cases of West Iranian {\it l} may be secondary rather than archaic \citep[262ff.]{Hubschmann1895}. 
It is not clear, for example, where forms like S.\ Tati (Ebrahim-abadi) {\it n\=albanda} $\sim$ (Sagz-abadi) {\it n\=arbanda} `elm' \citep[71]{Yarshater1969} $=$ NP {\it n\=arvan} belong.\footnote{The elm appears to be the frequent target of folk etymology in Iranian languages \citep[70]{Henning1963}; it is possible that Tati speakers have conflated the tree's name with {\it n\=al-band} `smith, farrier' (Arabic {\it na`l} `horseshoe'), on the basis of some perceived but non-obvious connection to horseshoes.} 
Similarly, one finds S.\ Tati {\it kelma} `worm' $=$ NP {\it kirm}; S.\ Tati {\it anjila} \citep[71]{Yarshater1969}, Vidari {\it in\v{\j}il} \citep[36]{Baghbidi2005} $=$ NP {\it anj\={\i}r} `fig' (forms elsewhere in Iranian point to \textit{*r}, e.g., Sogdian {\it an\v{c}\=er}, {\it anj\=er}; \citealt[37]{Qarib1995}). For `worm', the evidence clearly points to an Indo-Iranian etymon \textit{*kr(i)mi-} containing {\it r}, 
and any instances of {\it l} in Iranian languages should be secondary (e.g., Ossetic {\it k{\ae}lm} shows expected \textit{*r} $>$ {\it l} change in anticipation of \textit{*i} or \textit{*\textsubarch{i}}). 
Likely innovations are also found in Kurdish {\it valg}, Judeo-Tati {\it velg} \citep[]{Miller1892}, etc.\ $=$ NP {\it barg} $<$ \textit{*\textsubarch{u}arka-} \citep[47]{Horn1893}; this variant surfaces in the Dari dialect of New Persian as {\it balg} (\citealt[160]{Korn2005}). 
Non-archaic {\it l} can also be found in NP {\it \v{s}ik\=ar} `hunt' vs.\ Bandari, Bakhtiyari {\it e\v{s}k\=al}, S.\ Bashkardi {\it \v{s}ek\={a}l} `mountain sheep', if from a verbal root \textit{*skar-} with no good Indo-European cognates \citep[346]{Cheung2007}. 
Ultimately, {\it r $\sim$ l} variation across West Iranian is due not only to preservation of original PIE \textit{*l}, but also a secondary change to {\it l} from original \textit{*r}, especially evident in loans originally from non-Iranian languages, e.g., Judeo-Isfahani {\it kelews} `celery' $=$ NP {\it karafs} \citep{Stilo2007} $\leftarrow$ Arabic. 
We can be sure of the directionality in cases where there is secure evidence from outside of Indo-Iranian, but in the absence of such information, it can be difficult to tease apart primary and secondary {\it l}; it is equally unclear whether all variant pronunciations stem from the same dialectal source.


\subsection{Changes to PIr \textit{*\textsubarch{u}-}}
\label{W}
Reflexes of PIr \textit{*\textsubarch{u}-} are characterized by a high degree of irregularity across West Iranian.\footnote{See \citealt{Schwartz1982} on certain conditioned reflexes of this sound.} 
Developments within Persian serve to demonstrate the complexity of these developments. 
Proto-Iranian \textit{*\textsubarch{u}-} surfaces as Middle Persian {\it g-} before \textit{*\textsubring{r}} and \textit{*\textsubarch{i}}, but is otherwise unchanged in Middle Persian (with a few stray exceptions; see below):
\begin{description}[noitemsep]
\item PIr \textit{*\e{\textsubarch{u}}\textsubring{r}tka-} $>$ MP {\it \e{g}urdag} $>$ NP {\it \e{g}urdah} `kidney'
\item PIr \textit{*\e{\textsubarch{u}}\textsubring{r}ka-} $>$ MP {\it \e{g}urg} $>$ NP {\it \e{g}urg} `wolf'
\item PIr \textit{*\e{\textsubarch{u}}\textsubring{r}pa-ka-} $>$ MP {\it \e{g}urbag} $>$ NP {\it \e{g}urbah} `cat' (cf.\ YAv {\it urupi-} `dog, fox (?)' $<$ \textit{*\textsubarch{u}\textsubring{r}pi-})
\item PIr \textit{*\e{\textsubarch{u}}(a)\textsubring{r}da-}\footnote{YAv {\it var{\textschwa}$\delta$a-} `rose' points to (and Semn {\it val(a)} `flower' seems to point to --- perhaps also Pth {\it w'r} {\IPA /wa:r/}) \textit{*\textsubarch{u}arda-}, while the Persian forms, along with Gor {\it wil\={\i}}, may point to \textit{*\textsubarch{u}\textsubring{r}da-} (\citealt[77]{Mackenzie1961} gives the former etymon for all these forms).} $>$ MP {\it \e{g}ul} $>$ NP {\it \e{g}ul} `flower, rose'
\item PIr \textit{*\e{\textsubarch{u}}\textsubring{r}\v{s}na-ka-}\footnote{See \citealt[208]{Klingenschmitt2000} regarding the reconstruction of this form (departing from the earlier reconstruction of \citealt[92]{Hubschmann1895}), as well as the double reflex {\it \v{s}} $\sim$ {\it r(V)s}. 
} `hungry' 
$>$ MP {\it \e{g}u\v{s}nag}, {\it \e{g}ursag}; 
NP {\it \e{g}u\v{s}nah}, {\it \e{g}urusnah}; 
Bakhtiyari {\it \e{g}osne}; Balochi (Marw) {\it \e{g}u\v{s}nag}; Gaz {\it \e{v}a\v{s}\v{s}e}; Larestani {\it \e{g}o\v{s}na}; Mazandarani {\it \e{v}a\v{s}n\=a}; Sivandi {\it \e{f}e\v{s}e} `qui a faim' (showing the Central Iranian development \textit{*\textsubarch{u}-} $>$ {\it f-}, \citealt[cf.][]{Asatrian2012}); Taleshi {\it \e{v}e\v{s}i}; S Tati {\it \e{g}o\v{s}na}; Zaz {\it \e{v}ey\v{s}\=an}
%
\item PIr \textit{*\e{\textsubarch{u}}\textsubarch{i}\={a}na-} $>$ OP \textit{*\e{v}iy\=ana} (?) $>$ MP {\it \e{g}y\=an} $>$ NP {\it \e{j}\=an} `life, soul'
\item PIr \textit{*\e{\textsubarch{u}}\textsubarch{i}\={a}ka-} $>$ OP \textit{*\e{v}iy\=aka} (?) $>$ MP {\it \e{g}y\=ag} $>$ NP {\it \e{j}\=ah} `place'
\end{description}
Generally speaking, PIr \textit{*\textsubarch{u}\u{\i}-} $>$ MP {\it wi-} $>$ NP {\it gu-}:
\begin{description}[noitemsep]
\item PIr \textit{*\e{\textsubarch{u}}i-n\=a\'ca-} $>$ MP {\it \e{w}in\=ah} $>$ NP {\it \e{g}un\=ah} `sin'
\item PIr \textit{*\e{\textsubarch{u}}i-\v{c}\=ara-} $>$ MP {\it \e{w}iz\=ar} $>$ NP {\it \e{g}uz\=ar(dan)}
\item PIr \textit{*\e{\textsubarch{u}}i-d\=ana-} $>$ MP {\it \e{w}iy\=an} $>$ NP {\it \e{g}iy\=an} `tent' (cf.\ OInd {\it vi-dh\=a-} `furnish, spread, diffuse'?)
\item PIr \textit{*\e{\textsubarch{u}}a/in\v{\j}\=e\v{c}a-ka-} (?) $>$ MP {\it \e{w}in\v{\j}i\v{s}k} $>$ {\it \e{b}inji\v{s}k} $\sim$ {\it \e{g}unji\v{s}k} `sparrow', cf.\ Baxtiy\=ar\={\i} {\it \e{b}ingi\v{s}t} \citep[236]{Schapka1972}; cf.\ Challi {\it \e{v}e\v{s}kenj} \citep[69]{Yarshater1969}
\end{description}
However, some exceptions exist:
\begin{description}[noitemsep]
\item MP {\it \e{w}i\v{s}kar} $>$ NP {\it \e{b}i\v{s}gar(d)} `hunting ground'
\item MP {\it \e{w}iy\=ab\=an} `desert' $>$ NP {\it \e{b}iy\=ab\=an}
\end{description}
In the following forms, PIr \textit{*\textsubarch{u}V(C)r-} $>$ MP {\it wVr-} $>$ NP {\it gVr-}:
\begin{description}[noitemsep]
\item PIr \textit{*\e{\textsubarch{u}}ar\=a\'{\j}a-} $>$ MP {\it \e{w}ar\=az} $>$ NP {\it \e{g}ur\=az} `boar'
\item PIr \textit{*\e{\textsubarch{u}}art-} `turn' $>$ OP {\it \e{v}-r-t-} $>$ MP {\it \e{w}ardi\v{s}n} `turning' $>$ NP {\it \e{g}ardi\v{s}} (\citealt[87]{Mackenzie1971}; \citealt[423--5]{Cheung2007})
\item PIr \textit{*\e{\textsubarch{u}}a\'{\j}ra-} $>$ MP {\it \e{w}arz} $>$ NP {\it \e{g}urz} `mace' (cf.\ Bal {\it \e{b}urz} `club', \citealt[25]{Elfenbein1963})
\end{description}
Change to {\it g-} does not operate in the following words beginning with PIr \textit{*\textsubarch{u}V(C)r-}; most, but not all, have a grave (i.e., labial, labiodental, or velar) consonant later in the word:
\begin{description}[noitemsep]
\item PIr \textit{*\e{\textsubarch{u}}arna-ka-} $>$ MP {\it \e{w}arrag} $>$ NP {\it \e{b}arra} `lamb'
\item PIr \textit{*\e{\textsubarch{u}}a\'{\j}r(a)-ka-} $>$ OP {\it \e{v}-z-r-k} {\IPA /vaz\textsubring{r}ka-/}\footnote{\citet[69]{Schmitt1989} and others give this reading, departing from {\it vazraka-} (found in \citealt{Kent1953}), on the basis of the later forms. 
} $>$ Phl, MMP {\it \e{w}uzurg} (cf.\ Pazand {\it \e{g}uzurg}, \citealt[56]{Bailey1933}) $>$ NP {\it \e{b}uzurg}
\item PIr \textit{*\e{\textsubarch{u}}arka-} $>$ MP {\it \e{w}arg} $\sim$ {\it \e{w}alg} $>$ NP {\it \e{b}arg} `leaf'
\item PIr \textit{*\e{\textsubarch{u}}afra-} $>$ MP {\it \e{w}afr} `snow' $>$ \textit{*{\textbeta}afr-} (cf. Judeo-Persian <b\={p}r> {\IPA [bafr]}, \citealt[50]{Paul2013}) $>$ NP {\it \e{b}arf}
\item PIr \textit{*\e{\textsubarch{u}}ara-} $>$ MP {\it \e{w}ar} `breast' $>$ NP {\it \e{b}ar}
\item PIr \textit{*\e{\textsubarch{u}}ar-ma-} \citep[?\ cf.][298]{Horn1893} $>$ MP {\it \e{w}arm} `pond' $>$ NP {\it \e{b}arm}
\item MP {\it \e{w}arm} `memory' $>$ NP {\it \e{b}arm}
\item MP {\it \e{w}ardag} `captive, prisoner' $>$ NP {\it \e{b}arda}
\item PIr \textit{*\e{\textsubarch{u}}ar\'{\j}a-} $>$ MP {\it \e{w}arz} `work, agriculture' $>$ NP {\it \e{b}arz} `a sown field, agriculture' (cf.\ NP {\it varz\={\i}dan} `sow a field', with {\it v-})
\end{description}

Elsewhere, PIr \textit{*\textsubarch{u}-} $>$ MP {\it w-} $>$ NP {\it b-}:
\begin{description}[noitemsep]
\item PIr \textit{*\e{\textsubarch{u}}ah\=ana-ka-} \citep{Gershevitch1952} $>$ MP {\it \e{w}ih\=anag} (Phl <\e{w}h'n(k)>, MMP <\e{w}h'n(g)>) $>$ NP {\it \e{b}ah\=ana} `reason, pretext'
\item PIr \textit{*\e{\textsubarch{u}}ah\=ara-} $>$ MP {\it \e{w}ah\=ar} $>$ NP {\it \e{b}ah\=ar} `spring'
\item PIr \textit{*\e{\textsubarch{u}}ata-} $>$ NP {\it \e{w}ad} $>$ NP {\it \e{b}ad} `bad'
\item PIr \textit{*\e{\textsubarch{u}}at-\v{c}aka-} $>$ MP {\it \e{w}a\v{c}\v{c}ag} $>$ NP {\it \e{b}a\v{c}\v{c}ah} `child'
\item PIr \textit{*\e{\textsubarch{u}}ana-} $>$ MP {\it \e{w}an} $>$ NP {\it \e{b}un} `tree'
\item PIr \textit{*\e{\textsubarch{u}}\=ata-} $>$ MP {\it \e{w}\=ad} $>$ NP {\it \e{b}\=ad} `wind'
\item PIr \textit{*\e{\textsubarch{u}}\={\i}\'cati-} $>$ MP {\it w\={\i}st} $>$ NP {\it \e{b}\={\i}st} `twenty' (note also {\it s} for PIr \textit{*\'c}, while other decads show {\it h}) 
\item PIr \textit{*\e{\textsubarch{u}}ah\textsubarch{i}a-} $>$ Phl {\it \e{w}eh}, MMP {\it \e{w}eh}, {\it \e{w}ahy} $>$ NP {\it \e{b}ih} `better, good'
\item PIr \textit{*\e{\textsubarch{u}}a\'c\textsubarch{i}a-} $>$ MP {\it \e{w}\=e\v{s}} $>$ NP {\it \e{b}\=e\v{s}}  `more'
\item PIr \textit{*\e{\textsubarch{u}}ahi\v{s}ta-} `best' $>$ MP {\it \e{w}ahi\v{s}t} $>$ NP {\it \e{b}ihi\v{s}t}; cf.\ the name of a 4th cent.\ CE Christian martyr, {\it Gu(hi)\v{s}t\=az\=ad} \citep{Peeters1910}, the first member of which $<$ \textit{*\textsubarch{u}ahi\v{s}ta-}
\item PIr \textit{*\e{\textsubarch{u}}a/in\v{\j}\=e\v{c}a-ka-} (?) $>$ 
NP {\it \e{b}inji\v{s}k} $\sim$ {\it \e{g}unji\v{s}k} `sparrow' (see above) 
\item PIr \textit{*\e{\textsubarch{u}}r\twoacc[\u|\={\i}]n\v{\j}i-} $>$ MP {\it \e{b}rin\v{\j}} $>$ NP {\it \e{b}irinj} $\sim$ {\it \e{g}urinj} `rice' (cf.\ AV$+$ {\it vr\={\i}h\'i-})
\end{description}
As is apparent, none of the sound laws sketched above is exceptionless. It is almost certain that contact between closely related dialects is responsible for some of the doublets seen above. 
But it is also clear that succinct generalizations regarding the behavior of PIr \textit{*\textsubarch{u}-} in different conditioning environments are hard to come by. 
This issue has not received a systematic treatment in the literature. 
\citet[280--1]{Lentz1926} seems to consider \textit{*\textsubarch{u}-} $>$ {\it b-} the regular Southwest Iranian outcome. 
\citet[76]{Mackenzie1971} takes the change \textit{*\textsubarch{u}-} $>$ {\it b-} as a feature shared by Persian and Northern and Central Kurdish dialects, whereas ``[i]n most other W.Ir dialects {\it w-} is little modified in this position,  while in Bal.\ it has developed into {\it g(w)-}.''

Attempts to establish the regular behavior of PIr \textit{*\textsubarch{u}-} for non-Persian West Iranian languages have proved as difficult as for Persian. 
Early Judeo-Persian records, thought to typify a link between Middle and Modern Persian, present an equally challenging picture \citep[35ff.]{Paul2013}. 
An errant strain of Middle Persian shows {\it g-} for expected {\it b-}, e.g., Pazand {\it guzurg} : NP {\it buzurg} \citep[56]{Bailey1933}. 
A large number of West Iranian languages leave \textit{*\textsubarch{u}-} more or less unmodified (surfacing as {\it v}, {\it w} or {\it f} but more importantly not merging with PIr \textit{*g-}, \textit{*b-}),  but forms with {\it g-} and {\it b-} still preponderate. 
For instance, while Zazaki usually shows {\it v-} (e.g., {\it v\=a} `wind'), the word for `blood' is {\it g\=un\={\i}} $<$ \textit{*\textsubarch{u}ahuni-}\footnote{The word for blood shows irregular historical phonology across west Iranian. NP {\it x\=un} (MP {\it x\=on}) has either undergone a metathesis between \textit{*\textsubarch{u}} and \textit{*h}, or was subject to the same irregular {\it x-}prothesis as MP {\it x\=ayag}, NP {\it x\=aya} `egg', {\it xirs} `bear'. Parthian has {\it guxn}, with unexpected {\it g-}; Sivandi has {\it f\={\i}n}.} \citep{Paul1998b}. 
South Tati {\it varga} `leaf' sits alongside {\it beh\=ar-} `spring' \citep[95,~103,~110]{Yarshater1969}. 
The Kurmanji dialect of Kurdish shows a preference for {\it b-} where other languages do not, e.g., {\it bur\=az}, {\it vur\=az} `boar' : NP {\it gur\=az}; {\it birs\={\i}}, 
{\it bir\v{c}\={\i}} 
`hungry' : NP {\it gurusnah} \citep{Soane1913,Thackstonnd,Chyet2003}, but elsewhere agrees with Persian, e.g., {\it gurg}, {\it g\=ur} `wolf'.

If a regular outcome can be established for a given non-Persian language, there is a tendency to assume that any words containing deviations from it are loans from Persian (though this approach is in general avoided by \citealt{Korn2005}). 
For instance, Marw Balochi {\it burz} `mace' ($<$ \textit{*\textsubarch{u}a\'{\j}ra-}; note the metathesis identical to Persian) does not show expected {\it g(w)-}, hence, \citet[25]{Elfenbein1963} marks it as a ``Persic'' loan. 
However, there is no reason to expect NP {\it b-} in a reflex of a Middle Persian word with an initial syllable of the shape \textit{*\textsubarch{u}ar(C)-}, unless a grave consonant is found later in the word (and if the sound law sketched above is accurate). 
The Northern Kurdish dialect Kurmanji does, as mentioned above; this behavior can be found sporadically in other non-Persian languages as well (e.g., Mamasani {\it bur\'{\^{a}}z} `wild pig', \citealt[184]{MannHadank1909}). 
Given this evidence, these languages may be more viable donors for Balochi {\it burz} than Persian (the metathesis found in both of the forms is another question entirely). 

\subsection{Metathesis}
\label{meta}
Over the course of Persian history, more than one development of metathesis has taken place \citep[266--7]{Hubschmann1895}, involving the re-sequencing of word-final and some word-internal clusters ending in {\it r} (and on occasion {\it l}). 
By the advent of Middle Persian, we see {\it na\e{rm}} `soft' $<$ \textit{*na\e{mr}a-} and {\it wa\e{rz}} `club, mace' $<$ \textit{*\textsubarch{u}a\e{\'{\j}r}a-}. 
Fricative $+$ {\it r/l} clusters (as well as some fricative $+$ fricative clusters) have undergone metathesis after Middle Persian attestations:
\begin{description}[noitemsep]
\item PIr \textit{*\textsubarch{u}a\e{fr}a-} `snow' $>$ MP {\it wa\e{fr}} $>$ NP {\it ba\e{rf}}
\item PIr \textit{*ta\e{xr}a-} `bitter' $>$ Phl {\it ta\e{xl}}, MMP {\it ta\e{hr}} $>$ NP {\it ta\e{lx}}; Phl {\it ta\e{xl}\={\i}h} `bitterness' $>$ NP {\it ta\e{lx}\={\i}} (the latter change could be analogical)
\item PIr \textit{*\v{c}a\e{xr}a-} `wheel' $>$ MP {\it \v{c}a\e{xr}} $>$ NP {\it \v{c}a\e{rx}}
\item PIr \textit{*a\e{\'cr}u-} `tear' $>$ Phl {\it a\e{rs}}, MMP {\it a\e{rs}}, {\it a\e{sr}} $>$ NP {\it a\e{rs}} (alongside {\it a\e{\v{s}}k} $<$ \textit{*a\e{\'cr}u-ka})
\end{description}
Other West Iranian languages vary as to whether they show metathesis in the same words; this variation is often language internal:
\begin{description}[noitemsep]
\item PIr \textit{*\textsubarch{u}a\e{fr}a-} $>$ Bal {\it ba\e{rp}}; Gaz {\it va\e{f}}, {\it -va\e{rf}} (in compounds); Gur {\it va\e{rw}a}; Khun {\it va\e{rf}}; Lar {\it va\e{fr}}, {\it ba\e{rf}}; Maz {\it va\e{rf}}; Siv {\it va\e{rf}}; Tal {\it va\e{r}}; S Tati {\it va\e{r}a}; Zaz {\it ve\e{wr}}; Judeo-Tati {\it v\"a\e{hr}} `snow' can be found in the materials of \citet[59]{Miller1892}, but \citet[323]{Authier2012} gives {\it ve\e{rf}}. 
\item PIr \textit{*ta\e{xr}a-} $>$ Bal (Rakhshani) {\it ta\e{(h)l}}
\end{description}
Language contact must have played a role in bringing about intense variation, but the exact mechanisms are unclear. Metathesis is generally associated with Persian, since it can be documented in Persian's history. However, it is not clear whether the presence of metathesis in a non-Persian language is due to wholesale lexical borrowing or lexical diffusion (i.e., the adoption of the pronunciation {\it rC} for earlier {\it Cr}). 
Lexical borrowing from Persian tends to be assumed in the literature. 
For languages with {\it varf} : NP {\it barf}, it is assumed that the loan is from Middle Persian, or some period predating the change of MP {\it w-} to NP {\it b-}; for instance, \citet[749]{EilersSchapka1978} derives Gazi {\it v\"{a}rf} `snow' from MP {\it varf} [{\it sic}].
However, this is unlikely to be the case. If we take Judeo-Persian to be representative of the link between Middle and New Persian \citep[cf.][]{MacKenzie2003}, then Judeo-Persian forms like {\it <b\={p}r>} {\IPA [bafr]} \citep[50]{Paul2013} make it clear that metathesis postdates the merger of MP {\it w-} with {\it b-}, and that an intermediate stage \textit{*warf} was unlikely. 
Additionally, {\it w-}, {\it v-}, etc. cannot be secondary from earlier \textit{*b-} in the forms given above, since most of the languages mentioned show {\it b-} for original PIr \textit{*b-}.\footnote{Lenition often affects earlier intervocalic labial consonants, e.g., Qohrudi {\it v\={\i}x\=ov\=a} $=$ NP {\it b\=e-x\textsuperscript{v}\=ab} `sleeplessness' \citep[79]{Zhukovski1888}, NP {\it b\=e-} $<$ MP {\it ab\=e} $<$ \textit{*apa-ika-} \citep{DurkinMeisterernst2014}.} 

This detail aside, there are other reasons to question the account of lexical borrowing from Persian: 
first, this metathesis may not be a solely Persian development. Since most West Iranian languages (with exceptions, e.g., \citealt{Yarshater1962}) lost final syllable nuclei, it is likely that many languages had words ending in {\it -xr}, {\it -fr}, etc., clusters which posed articulatory and perceptual problems, and were resolved in a variety of ways, including metathesis. 
Second, many of the above forms can be analyzed only as {\it Mischformen}, vitiating a lexical borrowing account. 
Instead, it is possible that speakers in a situation of heavy multilingualism imposed pronunciations from forms in one language upon their cognates in another, a well-documented phenomenon in situations of multidialectalism, generally affecting less frequently uttered words \citep{Phillips1984,Stollenwerk1986,Wielingetal2011}.

\subsection{Changes affecting \textit{*dr}}

\citet[78--9]{Gershevitch1962locust} discusses reflexes of the word for `spade', demonstrating that some modern West Iranian languages reflect a form \textit{*barda-} (metathesized from \textit{*badra-}, which is internally derived from \textit{*badar-}). 
The source of metathesis in \textit{*barda-} is unclear. 
\citep[297--8]{Schwartz1970} shows that Iranian languages continue a doublet in the word for `grape', \textit{*angudra-} ($>$ MP, NP {\it ang\=ur}) $\sim$ \textit{*angurda-(ka-)} ($>$ NP {\it angurda}), the latter being secondary and a likely East Iranian loan into Persian and other languages. 
It is not clear whether the metathesis in \textit{*barda-} is a related phenomenon.\footnote{
\citeapos
{Bailey1973} derivation of the ethnonym Baloch from \textit{*ba$\delta$laut-\v{c}\={\i}} $<$ \textit{*\textsubarch{u}adra-\textsubarch{u}at(\v{c})\={\i}} `[land] having water [channels]' (cf.\ the Greek toponym {\it Gedrosia}) is criticized by \citet[47]{Korn2005} on the grounds that there is no parallel for \textit{*dr} $>$ \textit{*$\delta$l} $>$ {\it l}. However, this form may speak to a near-identical metathesis to \textit{*badra-}, \textit{*angudra-} etc., though the change \textit{*dr} $>$ \textit{*rd} $>$ {\it l} is a not a common Balochi development.}


\subsection{Prothetic \textit{x-}, \textit{h-}}
Two separate protheses have operated during the history of Persian. 
The first involves sporadic insertion of {\it x-} before an initial vowel, and predates Middle Persian; the second involves sporadic insertion of {\it h-} before an initial vowel, and predates New Persian.

These developments can be seen elsewhere in West Iranian, e.g., {\it xotk\=a} `duck' (language unmarked by \citealt[113]{Asatrian2012}) $<$ \textit{*\=ati-ka-}; Kumzari, Bandari, Larestani {\it xars} `tear' $<$ \textit{*a\'cru-} (cf.\ Bakhtiyari {\it hars}, Zazaki {\it hesri}). 
\citet[155--9]{Korn2005} provides a detailed treatment of this issue, 
and makes a strong case 
that some items showing initial {\it h-} in both Balochi and Kurdish are due to contact, though elsewhere, the sporadic presence of {\it h-} may be a sort of hypercorrection, as in many English dialects \citep[252--6]{Wells1982}, and not necessarily due to wholesale lexical borrowing (further bolstered by the fact that many Iranian languages lose initial {\it h-} under varying circumstances, e.g., \textit{*hi\'{\j}\textsubarch{u}\=ana-} `tongue' $>$ MP {\it izw\=an}, {\it uzw\=an}).

\subsection{\textit{\v{c}} $\sim$ \textit{\v{s}}}

Some quasi-systematic variation between {\it \v{s}} and {\it \v{c}} is found in forms across West Iranian. 
In some cases, original {\it \v{c}} becomes {\it \v{s}} due to the interference of Arabic, which lacks a phoneme {\it \v{c}} (in the relevant dialects), as in {\it \v{s}atranj} $\sim$ {\it \v{s}atrang} `chess' $<$ MP {\it \v{c}atrang} ($\leftarrow$ Old Indic {\it catur-a\.{n}ga-}). 

In other forms, as noted by \citet[71]{Horn1901}, {\it \v{c}} is secondary, e.g., Zor Yazdi {\it \v{c}\=um} `supper' $=$ NP {\it \v{s}\=am} `evening' (1st member $<$ \textit{*x\v{s}ap-}, cf.\ YAv {\it x\v{s}\=afniia-}, \citealt[553]{Bartholomae1904}); Kashani {\it \v{c}ilt\'uk} `unhulled rice' $=$ NP {\it \v{s}altok}; Kashani {\it cep\acm{u}n} `herdsman', Kurdish {\it \v{c}uw\=an} ({\it \v{c}\=op\=an} `butcher') $=$ NP {\it \v{c}ub\=an}, 
{\it \v{s}ub\=an} $<$ \textit{*f\v{s}u-p\=ana-} \citep{Horn1901}. 
Martin Schwartz (p.c.) 
points out that reflexes of the latter etymon may have undergone influence from NP {\it \v{c}\=ub} `staff, crook'.

\subsection{\textit{*t} $>$ \textit{r}}

The change \textit{*t} $>$ \textit{r} in North Tati dialects was noted by \citet[173]{Henning1954}. 
This change is seen in other languages, e.g., Judeo-Yazdi {\it \v{c}er-} `go' (< \textit{*\v{c}\textsubarch{i}uta-}), Judeo-Isfahani {\it \v{c}er-} `know' ($<$ \textit{*\v{c}ait-}), Kumzari {\it sp\={\i}r}, North Bashkardi {\it esp\={\i}r} `white' $<$ \textit{*\'c\textsubarch{u}aita-}. 
Some Central Dialects show variation between {\it \v{s}\={\i}r} $\sim$ {\it \v{s}\={\i}t} for `milk', though this may be due to the continuation of separate etyma \textit{*x\v{s}\={\i}ra-} and \textit{*x\v{s}\textsubarch{u}ifta-}.

\subsection{Other developments}

Above, a number of developments thought to be of interest to West Iranian dialectology were discussed. 
In this study, it is not possible to consider all possible meaningful changes, including vowel fronting \citep{Krahnke1976}, {\it p} $\sim$ {\it f} variation (e.g., S Tati {\it fercel} `dirty' : Bakhtiyari {\it par\v{c}al}), and other isoglosses. 
A hope is that as digitization efforts grow, fully data-driven approaches will allow us to take into account a wider range of innovations (see \S\ref{discussion} for details). 

\subsection{Key Issues}
\label{key.issue}

The foregoing sections served to illustrate the difficulties posed for the traditional comparative method by West Iranian sound change. 
Along the way, some problematic analytical decisions made by scholars have been highlighted, which are restated here:
\begin{itemize}
\item \citet{Elfenbein1963} assumes that Marw Balochi {\it burz} `mace' is a Persian loan, given unexpected {\it b-}, but it could easily be from another language (\S\ref{W})
\item \citet{EilersSchapka1978} assumes that Gazi {\it v\"arf} is a loan from Middle Persian \textit{*warf}, but no such form existed, given the relative chronology between the developments \textit{*\textsubarch{u}-} $>$ {\it b-} and \textit{*-fr} $>$ {\it rf}; if the metathesis shown by the Gazi form is due to Persian influence, lexical diffusion rather than lexical borrowing was likely involved (\S\ref{meta})
\item \citet{Korn2005} assumes that PIr \textit{*\'c\textsubarch{i}} $>$ Balochi {\it \v{s}} in all conditioning environments, and hence, that Balochi {\it k\=asib}/{\it kas\={\i}p} `turtle, tortoise', is a loan, but we cannot be sure this is the case (\S\ref{SY})
\end{itemize}

It is hoped that the qualitative points made or revived here --- namely that some of the segmental and prosodic contextual factors involved in West Iranian sound laws are indeterminate, that not all donor languages are necessarily Persian, and that pure lexical borrowing is not the sole mechanism of contact
 --- are convincing on their own merits. 
Still, it remains difficult to resolve many of the questions raised above within the constraints of the traditional comparative method. 
In general, it is difficult to maintain a bird's-eye view of the many innovations and archaisms that cut across the West Iranian lexicon; while discussing one type of variation, another type is ignored (the above discussion is no exception). 
The remainder of this paper develops a probabilistic methodology designed to relieve historical linguists of the need to make hard decisions regarding phonological outcomes in a dialectal group, and instead let regularities fall out of the data.






\section{Mixed Membership Models}
\label{HDP}

As described above, West Iranian languages show admixture from an unknown number of latent (i.e., unobserved or unknown) dialectal components, each with its own individual sound laws and analogical changes. 
The key aim of this work is to learn which underlying components have contributed various features to the noisy pattern observed. 
A number of statistical techniques exist for the purpose of reducing the dimensionality of multivariate categorical data; mixed-membership models of this sort learn clusters that capture co-occurrence patterns of features in a data set in a way that the human eye cannot easily manage to do. 
These include certain classes of so-called generative models, which attempt to tell a story specifying one or more latent parameters which are thought to have generated the observed data. 
The latent parameters specified in a generative model can be estimated, usually within a Bayesian framework, which infers their posterior distributions. 
Bayesian modeling allows prior distributions to be imposed over these parameters, which serves as a sometimes-necessary means of ensuring that the model embodies realistic behavior. 

\begin{figure}
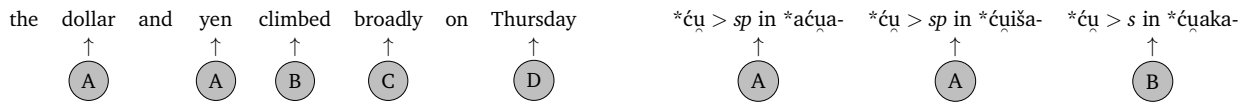

\begin{minipage}{.475\linewidth}
\begin{adjustbox}{max totalsize={\linewidth}{\linewidth},center}
\begin{tabular}{cccccccc}
the & dollar & and & yen & climbed & broadly & on & Thursday\\
 & $\uparrow$ &  & $\uparrow$ & $\uparrow$ & $\uparrow$ &  & $\uparrow$ \\
 & \tikz \node[draw,circle,fill=lightgray]{A}; &  & \tikz \node[draw,circle,fill=lightgray]{A}; & \tikz \node[draw,circle,fill=lightgray]{B}; & \tikz \node[draw,circle,fill=lightgray]{C}; &  & \tikz \node[draw,circle,fill=lightgray]{D}; \\
\end{tabular}
\end{adjustbox}
\end{minipage}
\hspace{.05\linewidth}
\begin{minipage}{.475\linewidth}
\begin{adjustbox}{max totalsize={\linewidth}{\linewidth},center}
\begin{tabular}{ccccc}
{*\'c\textsubarch{u} $>$ {\it sp} in *a\'c\textsubarch{u}a-} & {*\'c\textsubarch{u} $>$ {\it sp} in *\'c\textsubarch{u}i\v{s}a-} & {*\'c\textsubarch{u} $>$ {\it s} in *\'c\textsubarch{u}aka-}\\
$\uparrow$ & $\uparrow$ & $\uparrow$ \\
\tikz \node[draw,circle,fill=lightgray]{A}; & \tikz \node[draw,circle,fill=lightgray]{A}; & \tikz \node[draw,circle,fill=lightgray]{B};\\
\end{tabular}
\end{adjustbox}
\end{minipage}
\caption{A schematic comparison of topic modeling and the approach used in this paper. In types of topic modeling such as Latent Dirichlet Allocation (LDA), it is assumed that each content word instance in each document in a corpus is generated by a given ``topic,'' as represented by the shaded circles. These assignments are unknown a priori and must be inferred from the data. The example on the left provides hypothetical posterior topic assignments for a sentence fragment, with individual topics generating word instances from similar semantic fields or spheres of reference. The example on the right provides a hypothetical assignment of dialect components to sound changes operating in words found in a single Iranian speech variety.}
\label{topicmodel}
\end{figure}

I draw upon probabilistic models of document classification in order to motivate the model I use in this paper. 
{\sc Topic modeling}, which seeks to identify the topics present in a set of documents by associating the words found in them with one or more topics, is a well-known application for Bayesian mixed-membership models. 
{\sc Latent Dirichlet Allocation} (LDA) is one such model \citep{Bleietal2003}; it assumes a fixed number of topics. It assumes that there is an overall distribution over possible topics, that each document has a specific distribution over topics, and that each word in each document is distributed according to a particular topic. 
The posterior global distribution over topics, document-specific topic distributions, and word-specific topic associations can then be inferred; it should be noted that if the procedure is entirely unsupervised, topics will receive meaningless labels such as ``Topic 1'' rather than ``History,'' and that these labels require further interpretation. 
LDA is highly similar to the Structure algorithm of population genetics \citep{Pritchardetal2000}, which has been used in some linguistic applications \citep{Reesinketal2009,bowern2012riddle,longobardi2013toward,Syrjaenenetal2016}. 
Figure \ref{topicmodel} provides a hypothetical representation of how inferred topic assignments might appear when LDA is applied to a document classification as well as the data investigated in this paper.

LDA requires practitioners to provide a specification a priori of the total number of topics assumed. 
It is often unreasonable to assume that an exhaustive list of possible topics has been drawn up. 
LDA has a non-parametric extension, the {\sc Hierarchical Dirichlet Process} \citep[HDP,][]{Tehetal2005,Tehetal2006}, which allows for a potentially infinite number of topics. 
Over the course of the inference procedure, the model will return the number of topics which best explain the data. 

I wish to extend the HDP model to the problem of admixture in the vocabularies of Iranian languages. 
By aggregating the patterns of variation in reflexes of a number of Proto-Iranian etyma, we may be able to identify components in the lexicon of each language which conceivably can be explained via historical language contact. 
I assume that there exists a 
set of areal components which underlie the variation reflected synchronically in West Iranian languages, and that we can recover their associations with variants and representation within languages. 

An advantage of Bayesian models of this sort over classical methods for categorical data analysis 
is that they are generally robust to uneven or missing data --- this is critical, given the patchy coverage for some Iranian languages. 
At the same time, mixed-membership models can potentially be sensitive to skews in data coverage. If a large number of features bearing on a particular 
isogloss are well attested in the data, but others are not, the algorithm used to infer component distributions may learn a distribution based on the former, even when the latter are highly relevant (but under-attested).\footnote{A general practice is to remove uninformative and redundant features as well.} 
For this reason, I have taken pains to cast a wide net in the selection of features whilst maintaining parity in terms of the number of data points pertaining to each feature. 

\section{Feature selection and representation}
For the upcoming analysis, words exhibiting the relevant Proto-Iranian sounds and sound sequences were collected from grammars and dictionaries by searching for the relevant semantic field, yielding a dataset of 1229 words. 
It is acknowledged that this means of data collection is highly limited, as some languages are better etymologized than others, and it would be preferable to take a top-down approach to data collection using a digitized etymological dictionary or etymological database, when such resources are developed. 

As mentioned above, the goal here is to tease apart effects of areal contact and conditioning environments within West Iranian. 
As a concrete example, the presence of {\it b} in Sorani Kurdish {\it baran} ($<$ \textit{*\textsubarch{u}\=ar-}) versus {\it g} in Sorani Kurdish {\it gurg} ($<$ \textit{*\textsubarch{u}\textsubring{r}ka-}) is due either to contact (e.g., the language has taken the words over from different donor languages) or different conditioning environments in the two words triggering the changes \textit{*\textsubarch{u}} $>$ {\it b-} and \textit{*\textsubarch{u}} $>$ {\it g-}. 
Information regarding conditioning environments is key to the feature representation which serves as model input. 
However, explicitly stipulating conditioning environments requires too many assumptions. 
I use the {\sc etymon itself} as a proxy for conditioning environments; stating that \textit{*\textsubarch{u}} $>$ {\it b-} in the etymon \textit{*\textsubarch{u}\textsubring{r}ka-} {\sc wolf} is akin to stating that the change is triggered by the following \textit{*-\textsubring{r}-} and/or the following \textit{*-k-}. 
This would be a highly uneconomical analysis for a traditional historical grammar; 
however, any redundancy that this representation entails will be picked up by the model as part of the dimensionality reduction that it carries out.

A potential concern is that morphological variants of the same etymon are reflected in the catalogue of features; 
as mentioned above, different languages may continue different variants of a historical doublet \textit{*\textsubarch{u}\textsubring{r}da-/*\textsubarch{u}arda-} `flower'. 
A similar concern is that of homophony between reconstructed etyma, namely formally identical items that cannot be straightforwardly unified semantically (e.g., \textit{*\textsubarch{u}arma-} $>$ NP {\it barm} `pond, reservoir' and \textit{*\textsubarch{u}arma-} $>$ NP {\it barm} `memory'\footnote{These forms may be comprised by a single polysemous lexeme `pond, reservoir, memory', but are given separate headwords in \citealt{Mackenzie1971}; furthermore, colexification between `pond' and `memory' is not well attested cross-linguistically \citep{CLICS}.}). 
I leave the first problem untreated, with the hope that if a number of morphological variants of a single etymon are reflected in the data, this variation will be detectable in the model's output, namely via uncertainty in component level sound change distributions concerning this etymon.\footnote{A pervasive issue in the historical morphology of Indo-Iranian languages is the widespread use of the {\it *-aka-} suffix. The {\it k} of this suffix has been elided in most modern West Iranian languages, including New Persian \citep{Pisowicz1985}, making it difficult to determine whether certain forms in fact reflect {\it *-aka-}. In general, I do not make a distinction between suffixed and unsuffixed forms, unless there is clear widespread evidence for a suffix, as in the case of {\it a\v{s}k} `tear', found in New Persian, Gilaki, and other dialects.} 
I address the second problem by merging formally identical but semantically disparate reconstructions with one another, rather than treating them as instantiating different conditioning environments.



For the purposes of the model, each unobserved dialect component has a collection 
of {\sc sound change parameters} associated with it. I envision this to be a {\sc categorical} probability distribution over the {\sc possible observed outcomes} for each PIr sound of interest in each etymon (our proxy for the {\sc conditioning environment}). 
These parameters can be visualized as shown in Table \ref{hypothetical}, for a given dialect component (probabilities are hypothetical):

\begin{table}
\centering
\begin{tabular}{|l||lll|ll|}
\hline
{\sc etymon} & \textit{*\textsubarch{u}-} $>$ {\it b} & \textit{*\textsubarch{u}-} $>$ {\it g} & \textit{*\textsubarch{u}-} $>$ {\it w/v} & \textit{*\textsubring{r}\'{\j}} $>$ {\it Vl} & \textit{*\textsubring{r}\'{\j}} $>$ {\it Vrz}\\
\hline
\hline
\textit{*\textsubarch{u}ar-} & .99 & .005 & .005 & \cellcolor{gray} & \cellcolor{gray} \\
\hline
\textit{*\textsubarch{u}\textsubring{r}ka-} & .005 & .005 & .99 & \cellcolor{gray} & \cellcolor{gray} \\
\hline
\textit{*sp\textsubring{r}\'{\j}an-} & \cellcolor{gray} & \cellcolor{gray} & \cellcolor{gray} & .99 & .01\\
\hline
\textit{*\textsubring{r}\'{\j}if\textsubarch{i}a-} & \cellcolor{gray} & \cellcolor{gray} & \cellcolor{gray} & .97 & .03\\
\hline
\end{tabular}
\caption{Hypothetical sound change probabilities for a latent dialect component. Note that probabilities of outcomes for the relevant PIr sound(s) sum to one, and that distributions are {\sc sparse} (with the majority of mass concentrated on one outcome).}
\label{hypothetical}
\end{table}

Under the Neogrammarian hypothesis, sound change is exceptionless \citep{OsthoffBrugmann1879,Bloomfield1933,Hoenigswald1965,MorpurgoDavies1978}. 
The probability of a sound change operating in a given speech variety is strictly categorical: one outcome will occur with 100\% probability, all others with 0\% probability. 
This paper's model {\sc relaxes} the Neogrammarian hypothesis, allowing sound change probabilities to be non-categorical. 
The first purpose is practical: rigid categorical-valued variables which assign zero, rather than infinitesimal probability mass to an outcome, will cause problems for the inference procedure, and enumerating all possible combinations of categorical feature states is computationally unfeasible. 
The second pertains to the real world, namely, to account for irregularity within a component that cannot be explained (due to analogy, so-called ``sporadic'' change, or some other mechanism). 
However, it is still ideal to constrain these probability distributions such that they are {\sc sparse}, with the majority of mass concentrated on one outcome, rather than {\sc smooth} (i.e., with mass distributed quasi-uniformly across outcomes). 
Ultimately, while we cannot constrain the model to enforce {\sc regular} sound change, we can employ priors that {\sc regularize} sound change, encouraging probabilities to be very close to either 0 or 1. 

For the purposes of this study, I make no attempt to model intermediate stages in sound change. For instance, it is not entirely clear whether the {\it f-} in Sivandi {\it fin} `blood' $<$ \textit{*\textsubarch{u}ahuni-} comes from an intermediate \textit{*x(\textsubarch{u})}, or directly from \textit{*\textsubarch{u}} (though the latter scenario is more likely, as such changes are better attested in Sivandi). 
Techniques have been proposed for reconstructing forms at intermediate nodes on fixed phylogenies \citep{BouchardCoteetal2007,BouchardCoteetal2013}, but not for situations like ours, where a form in a given language is generated by one of an unknown number of dialect components, rather than a single fixed ancestor.\footnote{It is also worth noting the existence of the mStruct model \citep{ShringarpureXing2009}, a generalization of the Structure model which allows for mutations between historical and present-day populations. This model has the potential to account for intermediate stages, but situations of the sort described above are too rare in this paper's data set to justify the implementation of this considerably more complex methodology.} 
The relatively abstract model of feature representation employed at least partly ensures that the sound changes dealt with by the model are meaningful. 
This paper's data set comprises 1160 sound change instances instantiating 190 unique sound change types in 32 West Iranian languages. 







\section{Inference}
\label{inference}
The generative process underlying the HDP and the technical details of inference can be found in the appendix. 
A non-technical description of the HDP follows. 
Each data point (i.e., the reflex of a Proto-Iranian sound in a particular etymon in a given language, e.g., PIr \textit{*\textsubarch{u}-} $>$ NP {\it b-} in \textit{*\textsubarch{u}arma-}) is associated with a latent dialect component. 
The probability that a data point is associated with a given latent dialect component is dependent on a language-level probability distribution over dialect components $\boldsymbol \theta$, as well as a component-level distribution over sound changes $\boldsymbol \phi$. 
We do not know the values of these parameters, and must infer parameter values of high posterior probability (i.e., of high likelihood as well as high prior probability) from the data. 
Additionally, we do not know the true number of dialect components; this unknown must be learned by the model as well. 

The HDP involves three hyperparameters: 
$\alpha$ is the concentration parameter of the symmetric Dirichlet prior over each dialect component's {\sc sound change} distribution; 
the parameter $\gamma$ controls the dispersion of data points across dialect components within a given language; 
$\delta$ controls the number of components inferred (at the risk of oversimplifying). 
These hyperparameters 
can be fixed, or (as in the case of the parameters described in the previous paragraph) given a fully Bayesian treatment by estimating them from the data. 

Parameter and hyperparameter values can be estimated in several ways, including Markov chain Monte Carlo (MCMC) approaches such as Gibbs Sampling \citep{GemanGeman1984} or Variational Bayesian methods \citep{Bishop2006}. 
In the former procedure, values for each parameter are sampled stochastically on the basis of current values of all other parameters; after many iterations, the Gibbs sampler is guaranteed to draw samples from the posterior distribution of each parameter. 
Variational methods can be either deterministic or stochastic, and unlike MCMC methods, they assume a parametric form of the posterior distribution of each variable known as the variational posterior distribution, the parameters of which are iteratively updated. 
I use Automatic Differentiation Variation Inference \citep[ADVI, ][]{Kucukelbiretal2017}, as implemented in PyMC3 \citep{Salvatieretal2016} to infer the posterior distributions of $\boldsymbol \theta$ and $\boldsymbol \phi$ (as described in the Appendix). 






\section{Results}
\label{results}
As stated in the previous section, the inference procedure finds posterior probability distributions for two key parameters: $\boldsymbol \theta$, which gives each language's posterior distribution over dialect components;  $\boldsymbol \phi$, which gives each dialect component's distribution over sound changes. 

\begin{figure}[h!]
\centering
\begin{adjustbox}{width=.9\linewidth}
\input{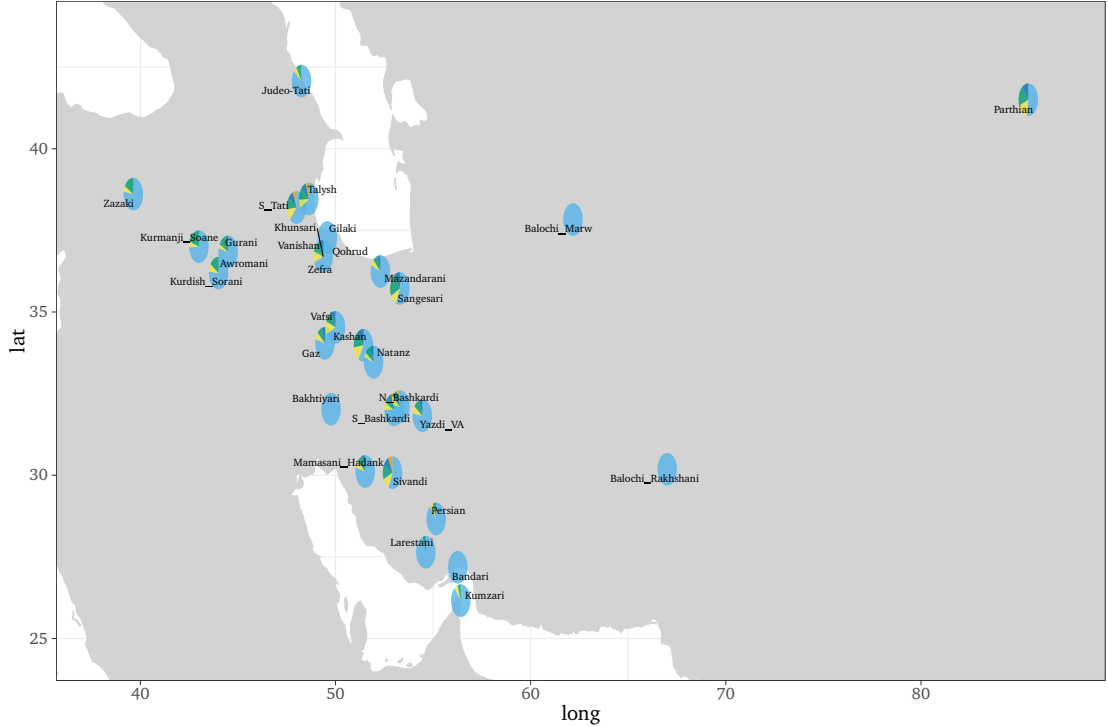}
\end{adjustbox}
\caption{Language-level posterior distributions over latent dialect components}
\label{comp.map}
\end{figure}

\subsection{Language-level component distributions}
As is clear from Figure \ref{comp.map}, most languages in the sample show a relatively uniform profile in terms of their component makeup, favoring a small number of identical components. 
This pattern dovetails with received wisdom regarding the widespread dominance of Persian over other West Iranian languages in the period following the Safavid empire roughly 500 years before the present day \citep{Borjian2009}; this homogenization appears to have resulted in a more or less uniform profile for New West Iranian languages in terms of the sound changes reflected in their vocabularies (albeit with some degree of differentiation). 

Virtually all languages in the sample show some degree of admixture from component $k=1$, along with differing degrees of components $k\in\{2,3,4\}$. 
Interestingly, $k=1$ appears to be strongly associated with developments that are thought to be typical of NW Iranian, such as the retention of initial {\it *\textsubarch{u}} and the non-operation of the change {\it *$\theta$r} $>$ {\it s}. It is not surprising that Modern Persian attests this component to a strong degree, given the well-known NW Iranian component in its vocabulary; at the same time, this proportion is higher than expected, as certain instances of SW Iranian behavior are strongly associated with this component (e.g., {\it \v{s}t $>$ st} in {\it *mu\v{s}ti-} fist, with {\it \v{s}t $>$ \v{s}t} receiving higher posterior probability in components $k\in\{2,3\}$. 

While visualizing  
$\boldsymbol \theta$ 
gives us an overview of the predominant components present in a language's vocabulary, the picture presented is difficult to interpret in that it does not allow us to pinpoint exactly why 
these components are present, in terms of the reflexes to which they are linked. To gain a closer understanding of this issue, it is instructive to inspect 
the posterior probabilities of component membership for 
individual sound change instances. 
I describe these issues in detail in the upcoming section.

\subsection{Posterior distributions over components for sound change instances}
I use the MAP values of $\boldsymbol \theta, \boldsymbol \phi$ to reconstruct the posterior probability distribution over component membership for each individual token with index $i$ --- i.e., each sound change instance in each language --- in the data set, $P(z_i|\boldsymbol \theta, \boldsymbol \phi)$. 
These probability distributions 
are given in the Appendix, as well as a table summarizing these values by averaging them across instances for each sound change type. 
These values allow us to address hypotheses about the provenance of certain sound changes (such as those discussed in \S \ref{key.issue}). 
Many of these distributions exhibit high uncertainty or entropy, with probability mass spread out across more than one dialect group rather than concentrated on a single group; this is perhaps a consequence of the relatively small size of the data set used in this study. 
At first glance, this uncertainty may seem to make the results difficult to interpret, but on the contrary these results are quite interpretable in that this uncertainty is relatively informative. 
Consider the following posterior distributions, concerning reflexes of Proto-Iranian {\it *b\textsubring{r}\'{\j}ant-} `high' and {\it *\'{c}\textsubarch{u}aka-} `dog', which show the posterior probability of a sound change type given a dialect component (I exclude components where none of the probabilities exceed $.05$ for visual clarity).
\begin{center}
{\tiny 
\begin{tabular}{lllllllll}
etymon & sound & reflex & $p(k=1)$ & $p(k=2)$ & $p(k=3)$ & $p(k=4)$ & $p(k=5)$ & $p(k=6)$\\
\hline
b\textsubring{r}\'{\j}ant & r\'{\j} & l & 0.58 & 0.11 & 0.15 & 0.08 & 0.04 & 0.02\\
b\textsubring{r}\'{\j}ant & r\'{\j} & r\'{\j} & 1.00 & 0.00 & 0.00 & 0.00 & 0.00 & 0.00\\
\hline
\'{c}\textsubarch{u}aka & \'{c}\textsubarch{u} & s & 0.01 & 0.23 & 0.34 & 0.16 & 0.09 & 0.10\\
\'{c}\textsubarch{u}aka & \'{c}\textsubarch{u} & sp & 1.00 & 0.00 & 0.00 & 0.00 & 0.00 & 0.00\\
\hline
\end{tabular}
}
\end{center}
\noindent Tokens exhibiting the change {\it *r\'{\j}} $>$ {\it r\'{\j}} (our shorthand for forms such as {\it burz}, which do not undergo change to {\it l}) are associated strongly with a single latent dialect component, $k=1$, as are tokens exhibiting the change {\it \'{c}\textsubarch{u}} $>$ {\it sp}. 
Tokens exhibiting the changes {\it *r\'{\j}} $>$ {\it l} and {\it *\'{c}\textsubarch{u}} $>$ {\it s} do not show a particularly strong affinity with any latent dialect component. 
What is critical here is that changes of the former type, usually associated with Northwest Iranian languages, show behavior that patterns much differently from changes usually associated with Southwest Iranian. This allows us to potentially classify individual change types according to whether the posterior distributions they exhibit are more in line with prototypical Northwest Iranian or Southwest Iranian sound changes. 

\begin{table}
\centering
{\tiny
\begin{tabular}{lllllllllllll}
etymon & sound & reflex & $p(k=1)$ & $p(k=2)$ & $p(k=3)$ & $p(k=4)$ & $p(k=5)$ & $p(k=6)$\\
\hline
\textsubarch{u}a\'{\j}ra & \textsubarch{u} & b & 0.08 & 0.17 & 0.39 & 0.17 & 0.07 & 0.02\\
\textsubarch{u}a\'{\j}ra & \textsubarch{u} & g & 1.00 & 0.00 & 0.00 & 0.00 & 0.00 & 0.00\\
\textsubarch{u}a\'{\j}ra & \textsubarch{u} & {$\gamma$} & 1.00 & 0.00 & 0.00 & 0.00 & 0.00 & 0.00\\
\hline
ka\'{c}\textsubarch{i}apa & \'{c}\textsubarch{i} & \v{s} & 0.03 & 0.24 & 0.28 & 0.15 & 0.06 & 0.18\\
ka\'{c}\textsubarch{i}apa & \'{c}\textsubarch{i} & s & 1.00 & 0.00 & 0.00 & 0.00 & 0.00 & 0.00\\
\hline
\textsubarch{u}afra & \textsubarch{u} & b & 0.70 & 0.06 & 0.12 & 0.04 & 0.03 & 0.03\\
\textsubarch{u}afra & \textsubarch{u} & w & 1.00 & 0.00 & 0.00 & 0.00 & 0.00 & 0.00\\
\textsubarch{u}afra & meta & meta & 0.01 & 0.25 & 0.34 & 0.18 & 0.08 & 0.07\\
\textsubarch{u}afra & meta & no meta & 1.00 & 0.00 & 0.00 & 0.00 & 0.00 & 0.00\\
\textsubarch{u}afra & meta & unclear & 1.00 & 0.00 & 0.00 & 0.00 & 0.00 & 0.00\\
\end{tabular}
}
\caption{Posterior component distributions for selected sound changes (I exclude components with probability mass under $.05$ for visual clarity)}
\label{change}
\end{table}

On the basis of these distributions, I propose provisional solutions for the problems identified in \S \ref{key.issue}. 
We find that \citeapos{Elfenbein1963} identification of Marw Balochi {\it burz} as a Southwest Iranian loan is indeed highly probable. Table \ref{change} shows the component distributions of changes affecting PIr {\it *\textsubarch{u}-} in {\it *\textsubarch{u}a\'{\j}ra-} `mace, club'. 
We see that change to {\it b-} shows a distribution similar to those of the prototypically Southwest Iranian sound changes discussed above, while change to {\it g-} and {\it $\gamma$-} shows Northwest Iranian behavior. 
Similarly, we find that change to {\it s} in {\it *ka\'{c}\textsubarch{i}apa-} `turtle/tortoise' patterns with canonically Northwest Iranian changes; hence, there is no strong reason to consider Balochi {\it k\=asib}/{\it kas\={\i}p} a loan, as assumed by \citet{Korn2005}, since it patterns with many other typically Balochi features. 
Finally, changes concerning the etymon {\it *\textsubarch{u}afra-} `snow' suggest a Northwest Iranian origin for the presence of {\it w-} and a Southwest Iranian origin for metathesis in the form; hence, Gazi {\it v\"arf} is probably a genuine {\it Mischform}, {\it pace} \citet{EilersSchapka1978}, stemming perhaps from a scenario where speakers in contact with a neighboring dialect exhibiting metathesis imposed this sound change on their inherited reflex of {\it *\textsubarch{u}afra-}. 

The results from the model are by no means the final word on these issues, and it is to be stressed that the conclusions drawn above are only tentative. 
It is likely that in many cases of idiosyncratic or unusual behavior, the paucity of data employed is the culprit. 
I have demonstrated however that this sort of methodology serves as a promising technique for teasing apart questions concerning dialectal admixture in Iranian and other dialect groups. 
I am confident that this method will produce increasingly realistic and reliable results as digital resources for Iranian languages grow, facilitating big data approaches to questions such as those addressed in this paper. 


\section{Discussion and future directions}
\label{discussion}
In this paper, I outlined a series of unresolved problems in Iranian dialectology and developed a probabilistic methodology designed to address these problems. 
In doing this, for the most part, I sought proof of concept as to whether Bayesian applications to Iranian dialectology might yield results which shed light on outstanding problems in the field as well as those that jibe with received wisdom. 
To some extent, this exercise was a success: I have shown that this model has great potential for resolving questions of the sort asked in this paper, but will benefit from further refinement. 
Below, I identify future directions that will improve this line of research:

\subsection{Data}
This paper made use of a relatively small data set compiled by hand from existing grammars. Sound changes were manually coded according to the behavior they displayed. Additionally, only sound changes thought to be of interest to West Iranian dialectology were included in the feature catalog. 
While I do not feel that this method of feature selection introduced any sort of pernicious bias that negatively affected results --- after all, this paper focused on patterns displayed by sound changes thought to be probative for the purposes of Iranian dialect grouping across the vocabularies of West Iranian languages --- it may be desirable to employ a more hands-off approach to feature selection and extraction, which will necessitate larger digitized etymological data sets. Additionally, this paper excluded East Iranian languages (including the languages Ormuri and Parachi), and shared patterns across both East and West Iranian should not be neglected; again, fulfilling this desideratum requires bigger data. 
At least two tacks can be taken for the purpose of data expansion: the first would involve digitizing of existing etymological dictionaries \citep{Cheung2007,RastorguevaEdelman} and converting them into a computationally tractable data format; however, no complete Iranian etymological dictionary currently exists for all parts of the lexicon, though current efforts such as the {\it Atlas of the Languages of Iran} \citep{Anonbyetal2019}, in its pilot phase at the time of writing, work towards filling this gap. 
The second approach involves applying semi-supervised cognate detection methods \citep{List2012,Rama2016} to digitized Iranian word lists, which can potentially be coupled with semi-supervised methodologies for linguistic reconstruction \citep{Melonietal2019}. While these methods still face many challenges, they can potentially save specialists a great deal of time and work in compiling large etymological resources. Whatever the approach employed, I believe that methods of the sort introduced in this paper will greatly benefit from the use of a larger data set. It is possible that the use of different data may yield different results from those reported in this paper. 

\subsection{Models}
While this paper employed the HDP, several alternative types of nonparametric mixed-membership model exist. 
The HDP has certain properties that are undesirable for certain uses, possibly including the dialectological application explored in this paper: 
specifically, the proportion of a component across all data points is correlated with its proportion within languages. 
It may be the case that a certain component is very rare overall, but well represented within one or a small number of languages. 
Certain alternatives to the HDP deal explicitly with this issue \citep{Williamsonetal2010}. 

\subsection{Representation of sound change}
In designing this paper's methodology, I made the radical decision to make no prior assumptions about the nature of the conditioning environments involved in the sound changes under study, instead treating entire etyma as conditioning environments. At first blush, this may seem like an implementation of the dictum that every word has its own history, attributed to dialect geographers such as Jules Gilli\'eron and Hugo Schuchardt. This is not the case: by linking the diachronic behavior of Proto-Iranian sounds in individual etyma to a finite number of dialect components exhibiting regularized sound change, we have inferred information regarding patterns of sound change within components as well as patterns of admixture within languages; the model ultimately embodies the interpretation of the above problem posed by dialect geographers that was provided by \citet[360]{Bloomfield1933}. 

At the same time, it may be wrong to ignore the effect of phonetic similarity between conditioning environments on sound change. 
It may be the case that in a particular dialect component, {\it *\textsubarch{u}-} undergoes a particular type of change in similar-looking etyma like \textit{*\textsubarch{u}ah\textsubarch{i}a-} and \textit{*\textsubarch{u}a\'c\textsubarch{i}a-}, but a different change in a more dissimilar etymon such as \textit{*\textsubarch{u}arka-}. 
I have ignored this possibility; my goal was to let this systematicity fall out of the data in a bottom-up fashion. 
If desired, it is possible to employ a prior over sound change that can express covariance, such as the logistic normal distribution, which will encourage Proto-Iranian sounds to behave similarly in phonetically similar environments (which can potentially be operationalized via a smooth kernel function of the edit distance between the etyma containing these environments).


\section{Conclusion}
This paper introduced a new way of looking at Iranian dialectal relationships. The focus was on sound change in West Iranian, but this method can potentially be extended to linguistic groups of similar geographic spread and time depth. 
My chief goal was to provide a means for relaxing assumptions regarding the operation of individual sound changes in individual languages, and allow regular patterns to fall out of the data. 
Much work remains to be done in order to understand the complex history of the Iranian languages. 
Larger data resources are needed, and cooperation among linguists is needed in order to design and refine the probabilistic models we use; as data analysts, we need to work together to characterize the stochastic processes that we believe to have generated the data we observe, formalized in probabilistic terms. 
There needs to be a willingness to simplify models (if particular models are intractable), and an effort to keep models flexible, so that they can be expanded. 
It is likely that many of these goals are well within reach. 

\section*{Acknowledgements}
All errors and infelicities are my own responsibility. Many of the issues treated in this paper are inspired by discussions with Martin Schwartz. I am additionally grateful for comments and suggestions provided by Tim Aufderheide, Florian Wandl, two anonymous referees, and editor James Clackson, as well as audiences at the Universities of Zurich and T\"ubingen.

\bibliographystyle{chicago}
\bibliography{iran_paper}

\section*{Appendix}

\subsection*{Model specification and inference}
The generative process for the HDP involving the truncated stick-breaking construction \citep{IshwaranJames2001} is given below. I set the truncation cutoff $T$, representing the maximum number of components, at $10$.\footnote{The choice of cutoff is arbitrary, and I have chosen a value I believe to be justified, given that the traditional literature hypothesizes the existence of a much smaller number of groups (around two or three), and allows for a degree of computational tractability (larger values lead to longer runtimes for inference).} $\mathcal{L}$ denotes the number of languages, $S$ the number of environments in which sound changes occur, and $N$ the number of data points in the data set. 
At a high level, this parameterization allows for the prior over components to be highly skewed such that certain components are favored and certain components have prior probabilities close to zero, as justified by the data.
\begin{description}[noitemsep]
\item Draw hyperparameters $\alpha,\delta,\gamma$, which control the sparsity of $\boldsymbol \phi$, dispersion of data points across components within languages, and the number of components inferred
\item $\boldsymbol \beta \sim \text{GEM}(\gamma)$ \hfill [draw atoms governing number of components learned]
\item $\boldsymbol \theta_{\ell,\cdot} \sim \text{Dirichlet}(\delta \boldsymbol \beta) : \ell \in \{1,...,\mathcal{L}\}$ \hfill [draw language-level distributions over component membership]
\item $\boldsymbol \phi_{t,s,\cdot} \sim \text{Dirichlet}(\alpha) : t \in \{1,...,T\}, s \in \{1,...,S\}$ \hfill [draw sound change parameters for every environment $s$ and every component $t$]
\item For $i \in \{1,...,N\}$ \hfill [for each data point (i.e., sound change instance)]
\begin{description}[noitemsep]
\item $z_i \sim \text{Categorical}(\boldsymbol \theta_{\ell_i,\cdot})$ \hfill [draw a component label]
\item $y_i|z_i = t \sim \text{Categorical}(\boldsymbol \phi_{t,x_i,\cdot})$ \hfill [sample the observed reflex on the basis of parameters associated with the label drawn in the previous step]
\end{description}
\end{description}
$\text{GEM}(\gamma)$ denotes the Griffiths-Engen-McCloskey distribution, which has the following function when parameterized by $\gamma$:

$$
\beta'_t \sim \text{Beta}(1,\gamma) : t \in \{1,...,T\}
$$
$$
\beta_t = \begin{cases} \beta'_t & \text{if}\ t =1\ \\ \beta'_t \prod_{i=1}^{t-1}(1 - \beta'_i) & \text{if}\ t \in \{2,...,T\} \end{cases}
$$
Under this process, each data point has the following likelihood:
$$
P(y_i,x_i,z_i=t|\boldsymbol \theta,\boldsymbol \phi) = P(z_i=t|\boldsymbol \theta_{\ell_i}) P(y_i|\boldsymbol \phi_{t,x_i})
$$
Marginalizing out the discrete variable $\boldsymbol z$ yields the following likelihood:
$$
P(y_i,x_i|\boldsymbol \theta,\boldsymbol \phi) = \sum_{t=1}^T P(z_i=t|\boldsymbol \theta_{\ell_i}) P(y_i|\boldsymbol \phi_{t,x_i})
$$
The posterior distributions of $\boldsymbol \theta,\boldsymbol \phi$ can be used to reconstruct the probability that a given data point is associated with a given dialect component:
$$
P(z_i=t|\boldsymbol \theta,\boldsymbol \phi, x_i, y_i) \propto \theta_{\ell_i,t} \phi_{t,x_i,y_i}
$$

I place uninformative $\text{Gamma}(1,1)$ priors over $\delta$ and $\gamma$, since we do not know {\it a priori} the degree to which data points within a given language should be dispersed across components, or how many components we should expect to find. 
I fix $\alpha$, the concentration parameter of the symmetric Dirichlet prior over each dialect component's {\sc sound change} distributions, at $.0001$ to encourage sparse sound change distributions. 

I carry out inference using ADVI \citep{Kucukelbiretal2017} in PyMC3 \citep{Salvatieretal2016}. 
ADVI allows users to define flexible and complex differentiable Bayesian models using a wide range of prior distributions over parameters. 
Certain probability distributions have constrained support: e.g., all samples from the Dirichlet distribution must be simplices summing to one; all samples from the Gamma distribution must be greater than zero. 
Parameters are 
mapped to unconstrained space and approximated with 
Gaussian variational posterior distributions, 
the parameters of which can be straightforwardly optimized using stochastic gradient descent. 
In mean-field ADVI, variational posteriors consist of independent Gaussian distributions, whereas in full-rank ADVI, variational posteriors make up a multivariate Gaussian distribution with non-diagonal covariance; I employ mean-field ADVI for simplicity. 
I optimize the model's variational parameters over 4 separate initializations of 100000 iterations each, monitoring the evidence lower bound (ELBO) for convergence. The learning rate and $\beta_1$ parameter of the Adam optimizer \citep{KingmaBa2014} are set to $.01$ and $.8$, respectively. Posterior samples for each parameter are generated by drawing 500 samples from the fitted variational posterior.

Mixture models suffer from the so-called label switching problem, in which indices of identical components differ across initializations/chains. To address this problem, I relabel the components inferred across initializations 2--4 by permuting component labels and selecting the permutation which minimizes the Kullback-Leibler divergence from the parameters for initialization 1 to the permuted parameters for the initialization under consideration. 
This allows us to average parameters across initializations, providing an approximation to the MAP configuration over component assignments for each item in the data set. 
Aggregating over these assignments produces MAP language-level distributions over component makeup. 

\subsection*{Posterior distributions over dialect components for sound change instances, averaged by type}
The following table gives $P(z_i|\boldsymbol \theta,\boldsymbol \phi, x_i, y_i)$ for every sound change instance in our data set, averaged across sound change {\sc types}. 
{\tiny
\begin{longtable}{lllllllllllll}
etymon & sound & reflex & $p(k=1)$ & $p(k=2)$ & $p(k=3)$ & $p(k=4)$ & $p(k=5)$ & $p(k=6)$ & $p(k=7)$ & $p(k=8)$ & $p(k=9)$ & $p(k=10)$\\
\hline
(f)\v{s}up\={a}na & \v{s} & \v{s} & 0.32 & 0.15 & 0.26 & 0.10 & 0.06 & 0.06 & 0.02 & 0.01 & 0.01 & 0.01\\
(f)\v{s}up\={a}na & \v{s} & c & 1.00 & 0.00 & 0.00 & 0.00 & 0.00 & 0.00 & 0.00 & 0.00 & 0.00 & 0.00\\
\hline
\'{\j}\textsubring{r}d & rd & l & 0.68 & 0.08 & 0.11 & 0.06 & 0.02 & 0.02 & 0.01 & 0.01 & 0.01 & 0.00\\
\'{\j}\textsubring{r}d & rd & rd & 0.92 & 0.00 & 0.00 & 0.00 & 0.08 & 0.00 & 0.00 & 0.00 & 0.00 & 0.00\\
\hline
\'{c}\textsubarch{i}\={a}\textsubarch{u}a & \'{c}\textsubarch{i} & \v{s} & 0.01 & 0.23 & 0.27 & 0.14 & 0.08 & 0.21 & 0.02 & 0.02 & 0.02 & 0.01\\
\'{c}\textsubarch{i}\={a}\textsubarch{u}a & \'{c}\textsubarch{i} & s & 1.00 & 0.00 & 0.00 & 0.00 & 0.00 & 0.00 & 0.00 & 0.00 & 0.00 & 0.00\\
\hline
\'{c}\textsubarch{u}aka & \'{c}\textsubarch{u} & s & 0.01 & 0.23 & 0.34 & 0.16 & 0.09 & 0.10 & 0.02 & 0.02 & 0.02 & 0.01\\
\'{c}\textsubarch{u}aka & \'{c}\textsubarch{u} & sp & 1.00 & 0.00 & 0.00 & 0.00 & 0.00 & 0.00 & 0.00 & 0.00 & 0.00 & 0.00\\
\hline
\'{c}\textsubarch{u}i\v{s}a & \'{c}\textsubarch{u} & s & 0.83 & 0.04 & 0.05 & 0.03 & 0.01 & 0.03 & 0.00 & 0.00 & 0.00 & 0.00\\
\'{c}\textsubarch{u}i\v{s}a & \'{c}\textsubarch{u} & sp & 1.00 & 0.00 & 0.00 & 0.00 & 0.00 & 0.00 & 0.00 & 0.00 & 0.00 & 0.00\\
\hline
\textsubarch{u}(a)rda & \textsubarch{u} & g & 0.01 & 0.22 & 0.35 & 0.16 & 0.08 & 0.10 & 0.02 & 0.02 & 0.02 & 0.02\\
\textsubarch{u}(a)rda & \textsubarch{u} & w & 1.00 & 0.00 & 0.00 & 0.00 & 0.00 & 0.00 & 0.00 & 0.00 & 0.00 & 0.00\\
\textsubarch{u}(a)rda & rd & l & 0.72 & 0.06 & 0.10 & 0.05 & 0.02 & 0.03 & 0.01 & 0.00 & 0.00 & 0.00\\
\textsubarch{u}(a)rda & rd & rd & 0.48 & 0.15 & 0.18 & 0.10 & 0.04 & 0.02 & 0.02 & 0.01 & 0.01 & 0.01\\
\hline
\textsubarch{u}\={a}\'{\j}i(\textsubarch{i}aka?) & \textsubarch{u} & b & 0.52 & 0.12 & 0.17 & 0.06 & 0.04 & 0.06 & 0.01 & 0.01 & 0.01 & 0.01\\
\textsubarch{u}\={a}\'{\j}i(\textsubarch{i}aka?) & \textsubarch{u} & g & 0.99 & 0.00 & 0.00 & 0.01 & 0.00 & 0.00 & 0.00 & 0.00 & 0.00 & 0.00\\
\textsubarch{u}\={a}\'{\j}i(\textsubarch{i}aka?) & \textsubarch{u} & w & 0.96 & 0.00 & 0.00 & 0.04 & 0.00 & 0.00 & 0.00 & 0.00 & 0.00 & 0.00\\
\hline
\textsubarch{u}\={a}\'{\j}i(n) & \textsubarch{u} & b & 0.10 & 0.22 & 0.34 & 0.15 & 0.08 & 0.05 & 0.02 & 0.02 & 0.01 & 0.01\\
\textsubarch{u}\={a}\'{\j}i(n) & \textsubarch{u} & w & 1.00 & 0.00 & 0.00 & 0.00 & 0.00 & 0.00 & 0.00 & 0.00 & 0.00 & 0.00\\
\hline
\textsubarch{u}\={a}f & \textsubarch{u} & b & 0.47 & 0.14 & 0.16 & 0.10 & 0.03 & 0.06 & 0.01 & 0.01 & 0.01 & 0.01\\
\textsubarch{u}\={a}f & \textsubarch{u} & g & 0.99 & 0.00 & 0.00 & 0.00 & 0.00 & 0.01 & 0.00 & 0.00 & 0.00 & 0.00\\
\textsubarch{u}\={a}f & \textsubarch{u} & w & 1.00 & 0.00 & 0.00 & 0.00 & 0.00 & 0.00 & 0.00 & 0.00 & 0.00 & 0.00\\
\hline
\textsubarch{u}\={a}r & \textsubarch{u} & b & 0.73 & 0.05 & 0.11 & 0.04 & 0.02 & 0.03 & 0.01 & 0.00 & 0.00 & 0.00\\
\textsubarch{u}\={a}r & \textsubarch{u} & g & 0.95 & 0.05 & 0.00 & 0.00 & 0.00 & 0.00 & 0.00 & 0.00 & 0.00 & 0.00\\
\textsubarch{u}\={a}r & \textsubarch{u} & w & 0.92 & 0.08 & 0.00 & 0.00 & 0.00 & 0.00 & 0.00 & 0.00 & 0.00 & 0.00\\
\hline
\textsubarch{u}\={a}ta & \textsubarch{u} & b & 0.63 & 0.08 & 0.14 & 0.05 & 0.03 & 0.04 & 0.01 & 0.01 & 0.01 & 0.00\\
\textsubarch{u}\={a}ta & \textsubarch{u} & g & 1.00 & 0.00 & 0.00 & 0.00 & 0.00 & 0.00 & 0.00 & 0.00 & 0.00 & 0.00\\
\textsubarch{u}\={a}ta & \textsubarch{u} & w & 1.00 & 0.00 & 0.00 & 0.00 & 0.00 & 0.00 & 0.00 & 0.00 & 0.00 & 0.00\\
\hline
\textsubarch{u}\textsubarch{i}\={a}ka & \textsubarch{u} & g & 0.67 & 0.05 & 0.18 & 0.03 & 0.04 & 0.02 & 0.00 & 0.00 & 0.00 & 0.00\\
\textsubarch{u}\textsubarch{i}\={a}ka & \textsubarch{u} & j & 0.79 & 0.05 & 0.08 & 0.04 & 0.02 & 0.02 & 0.00 & 0.00 & 0.00 & 0.00\\
\hline
\textsubarch{u}\textsubarch{i}\={a}na & \textsubarch{u} & g & 0.42 & 0.14 & 0.22 & 0.10 & 0.05 & 0.03 & 0.02 & 0.01 & 0.01 & 0.01\\
\textsubarch{u}\textsubarch{i}\={a}na & \textsubarch{u} & j & 1.00 & 0.00 & 0.00 & 0.00 & 0.00 & 0.00 & 0.00 & 0.00 & 0.00 & 0.00\\
\textsubarch{u}\textsubarch{i}\={a}na & \textsubarch{u} & y & 1.00 & 0.00 & 0.00 & 0.00 & 0.00 & 0.00 & 0.00 & 0.00 & 0.00 & 0.00\\
\hline
\textsubarch{u}\textsubring{r}\'{c}a-ka- & \textsubarch{u} & g & 0.12 & 0.18 & 0.31 & 0.15 & 0.09 & 0.07 & 0.02 & 0.02 & 0.02 & 0.02\\
\textsubarch{u}\textsubring{r}\'{c}a-ka- & r\'{c} & l & 0.24 & 0.16 & 0.27 & 0.13 & 0.08 & 0.06 & 0.02 & 0.02 & 0.01 & 0.02\\
\hline
\textsubarch{u}\textsubring{r}\v{s}naka & \textsubarch{u} & b & 0.01 & 0.18 & 0.49 & 0.10 & 0.10 & 0.07 & 0.01 & 0.01 & 0.01 & 0.01\\
\textsubarch{u}\textsubring{r}\v{s}naka & \textsubarch{u} & f & 1.00 & 0.00 & 0.00 & 0.00 & 0.00 & 0.00 & 0.00 & 0.00 & 0.00 & 0.00\\
\textsubarch{u}\textsubring{r}\v{s}naka & \textsubarch{u} & g & 1.00 & 0.00 & 0.00 & 0.00 & 0.00 & 0.00 & 0.00 & 0.00 & 0.00 & 0.00\\
\textsubarch{u}\textsubring{r}\v{s}naka & \textsubarch{u} & w & 1.00 & 0.00 & 0.00 & 0.00 & 0.00 & 0.00 & 0.00 & 0.00 & 0.00 & 0.00\\
\textsubarch{u}\textsubring{r}\v{s}naka & rs & \v{s} & 0.01 & 0.24 & 0.32 & 0.18 & 0.08 & 0.09 & 0.03 & 0.02 & 0.02 & 0.02\\
\textsubarch{u}\textsubring{r}\v{s}naka & rs & rs & 1.00 & 0.00 & 0.00 & 0.00 & 0.00 & 0.00 & 0.00 & 0.00 & 0.00 & 0.00\\
\hline
\textsubarch{u}\textsubring{r}ka & \textsubarch{u} & g & 0.34 & 0.16 & 0.23 & 0.11 & 0.06 & 0.06 & 0.02 & 0.01 & 0.01 & 0.01\\
\textsubarch{u}\textsubring{r}ka & \textsubarch{u} & w & 1.00 & 0.00 & 0.00 & 0.00 & 0.00 & 0.00 & 0.00 & 0.00 & 0.00 & 0.00\\
\hline
\textsubarch{u}\textsubring{r}tka & \textsubarch{u} & g & 0.02 & 0.24 & 0.34 & 0.16 & 0.08 & 0.11 & 0.02 & 0.02 & 0.02 & 0.01\\
\textsubarch{u}\textsubring{r}tka & \textsubarch{u} & w & 1.00 & 0.00 & 0.00 & 0.00 & 0.00 & 0.00 & 0.00 & 0.00 & 0.00 & 0.00\\
\hline
\textsubarch{u}a\'{\j}r(a)ka & \textsubarch{u} & b & 0.48 & 0.13 & 0.17 & 0.10 & 0.05 & 0.04 & 0.01 & 0.01 & 0.01 & 0.01\\
\textsubarch{u}a\'{\j}r(a)ka & \textsubarch{u} & g & 0.92 & 0.08 & 0.00 & 0.00 & 0.00 & 0.00 & 0.00 & 0.00 & 0.00 & 0.00\\
\textsubarch{u}a\'{\j}r(a)ka & \textsubarch{u} & w & 0.91 & 0.09 & 0.00 & 0.00 & 0.00 & 0.00 & 0.00 & 0.00 & 0.00 & 0.00\\
\hline
\textsubarch{u}a\'{\j}ra & \textsubarch{u} & b & 0.08 & 0.17 & 0.39 & 0.17 & 0.07 & 0.02 & 0.03 & 0.02 & 0.02 & 0.02\\
\textsubarch{u}a\'{\j}ra & \textsubarch{u} & g & 1.00 & 0.00 & 0.00 & 0.00 & 0.00 & 0.00 & 0.00 & 0.00 & 0.00 & 0.00\\
\textsubarch{u}a\'{\j}ra & \textsubarch{u} & {$\gamma$} & 1.00 & 0.00 & 0.00 & 0.00 & 0.00 & 0.00 & 0.00 & 0.00 & 0.00 & 0.00\\
\hline
\textsubarch{u}a\'{c}\textsubarch{i}a & \'{c}\textsubarch{i} & \v{s} & 0.71 & 0.08 & 0.09 & 0.06 & 0.02 & 0.02 & 0.01 & 0.01 & 0.00 & 0.00\\
\textsubarch{u}a\'{c}\textsubarch{i}a & \'{c}\textsubarch{i} & s & 0.64 & 0.12 & 0.11 & 0.07 & 0.03 & 0.01 & 0.01 & 0.01 & 0.01 & 0.00\\
\textsubarch{u}a\'{c}\textsubarch{i}a & \textsubarch{u} & b & 0.02 & 0.27 & 0.30 & 0.17 & 0.07 & 0.09 & 0.02 & 0.02 & 0.01 & 0.01\\
\textsubarch{u}a\'{c}\textsubarch{i}a & \textsubarch{u} & g & 1.00 & 0.00 & 0.00 & 0.00 & 0.00 & 0.00 & 0.00 & 0.00 & 0.00 & 0.00\\
\textsubarch{u}a\'{c}\textsubarch{i}a & \textsubarch{u} & w & 1.00 & 0.00 & 0.00 & 0.00 & 0.00 & 0.00 & 0.00 & 0.00 & 0.00 & 0.00\\
\hline
\textsubarch{u}afra & \textsubarch{u} & b & 0.70 & 0.06 & 0.12 & 0.04 & 0.03 & 0.03 & 0.00 & 0.00 & 0.00 & 0.00\\
\textsubarch{u}afra & \textsubarch{u} & w & 1.00 & 0.00 & 0.00 & 0.00 & 0.00 & 0.00 & 0.00 & 0.00 & 0.00 & 0.00\\
\textsubarch{u}afra & meta & meta & 0.01 & 0.25 & 0.34 & 0.18 & 0.08 & 0.07 & 0.02 & 0.02 & 0.02 & 0.01\\
\textsubarch{u}afra & meta & no meta & 1.00 & 0.00 & 0.00 & 0.00 & 0.00 & 0.00 & 0.00 & 0.00 & 0.00 & 0.00\\
\textsubarch{u}afra & meta & unclear & 1.00 & 0.00 & 0.00 & 0.00 & 0.00 & 0.00 & 0.00 & 0.00 & 0.00 & 0.00\\
\hline
\textsubarch{u}agza & \textsubarch{u} & b & 0.62 & 0.08 & 0.18 & 0.05 & 0.03 & 0.03 & 0.01 & 0.00 & 0.00 & 0.00\\
\textsubarch{u}agza & \textsubarch{u} & g & 0.98 & 0.00 & 0.00 & 0.00 & 0.01 & 0.01 & 0.00 & 0.00 & 0.00 & 0.00\\
\textsubarch{u}agza & \textsubarch{u} & m & 0.89 & 0.00 & 0.00 & 0.00 & 0.03 & 0.08 & 0.00 & 0.00 & 0.00 & 0.00\\
\textsubarch{u}agza & \textsubarch{u} & w & 0.96 & 0.00 & 0.00 & 0.00 & 0.01 & 0.03 & 0.00 & 0.00 & 0.00 & 0.00\\
\hline
\textsubarch{u}ah\={a}ra & \textsubarch{u} & b & 0.01 & 0.25 & 0.31 & 0.18 & 0.08 & 0.10 & 0.02 & 0.02 & 0.02 & 0.01\\
\textsubarch{u}ah\={a}ra & \textsubarch{u} & w & 1.00 & 0.00 & 0.00 & 0.00 & 0.00 & 0.00 & 0.00 & 0.00 & 0.00 & 0.00\\
\hline
\textsubarch{u}ah\textsubarch{i}a & \textsubarch{u} & b & 0.35 & 0.18 & 0.21 & 0.12 & 0.05 & 0.04 & 0.01 & 0.01 & 0.01 & 0.01\\
\textsubarch{u}ah\textsubarch{i}a & \textsubarch{u} & g & 1.00 & 0.00 & 0.00 & 0.00 & 0.00 & 0.00 & 0.00 & 0.00 & 0.00 & 0.00\\
\textsubarch{u}ah\textsubarch{i}a & \textsubarch{u} & w & 1.00 & 0.00 & 0.00 & 0.00 & 0.00 & 0.00 & 0.00 & 0.00 & 0.00 & 0.00\\
\hline
\textsubarch{u}ahuni & \textsubarch{u} & f & 0.00 & 0.24 & 0.30 & 0.19 & 0.05 & 0.15 & 0.02 & 0.02 & 0.02 & 0.02\\
\textsubarch{u}ahuni & \textsubarch{u} & g & 1.00 & 0.00 & 0.00 & 0.00 & 0.00 & 0.00 & 0.00 & 0.00 & 0.00 & 0.00\\
\textsubarch{u}ahuni & \textsubarch{u} & h & 1.00 & 0.00 & 0.00 & 0.00 & 0.00 & 0.00 & 0.00 & 0.00 & 0.00 & 0.00\\
\textsubarch{u}ahuni & \textsubarch{u} & w & 1.00 & 0.00 & 0.00 & 0.00 & 0.00 & 0.00 & 0.00 & 0.00 & 0.00 & 0.00\\
\textsubarch{u}ahuni & \textsubarch{u} & x & 1.00 & 0.00 & 0.00 & 0.00 & 0.00 & 0.00 & 0.00 & 0.00 & 0.00 & 0.00\\
\hline
\textsubarch{u}ai\v{s}a & \textsubarch{u} & b & 0.81 & 0.05 & 0.05 & 0.02 & 0.01 & 0.04 & 0.00 & 0.00 & 0.00 & 0.00\\
\textsubarch{u}ai\v{s}a & \textsubarch{u} & w & 0.72 & 0.07 & 0.09 & 0.05 & 0.02 & 0.02 & 0.01 & 0.01 & 0.01 & 0.00\\
\hline
\textsubarch{u}aic\={i}na? & \textsubarch{u} & b & 0.03 & 0.20 & 0.39 & 0.11 & 0.09 & 0.13 & 0.01 & 0.01 & 0.01 & 0.01\\
\textsubarch{u}aic\={i}na? & \textsubarch{u} & g & 1.00 & 0.00 & 0.00 & 0.00 & 0.00 & 0.00 & 0.00 & 0.00 & 0.00 & 0.00\\
\textsubarch{u}aic\={i}na? & \textsubarch{u} & w & 1.00 & 0.00 & 0.00 & 0.00 & 0.00 & 0.00 & 0.00 & 0.00 & 0.00 & 0.00\\
\hline
\textsubarch{u}aina & \textsubarch{u} & b & 0.02 & 0.24 & 0.33 & 0.14 & 0.07 & 0.14 & 0.02 & 0.01 & 0.01 & 0.01\\
\textsubarch{u}aina & \textsubarch{u} & g & 1.00 & 0.00 & 0.00 & 0.00 & 0.00 & 0.00 & 0.00 & 0.00 & 0.00 & 0.00\\
\textsubarch{u}aina & \textsubarch{u} & w & 1.00 & 0.00 & 0.00 & 0.00 & 0.00 & 0.00 & 0.00 & 0.00 & 0.00 & 0.00\\
\hline
\textsubarch{u}aita & \textsubarch{u} & b & 0.20 & 0.18 & 0.31 & 0.12 & 0.07 & 0.07 & 0.02 & 0.01 & 0.01 & 0.01\\
\textsubarch{u}aita & \textsubarch{u} & w & 1.00 & 0.00 & 0.00 & 0.00 & 0.00 & 0.00 & 0.00 & 0.00 & 0.00 & 0.00\\
\hline
\textsubarch{u}ana & \textsubarch{u} & b & 0.02 & 0.26 & 0.28 & 0.17 & 0.09 & 0.10 & 0.02 & 0.02 & 0.02 & 0.01\\
\textsubarch{u}ana & \textsubarch{u} & w & 1.00 & 0.00 & 0.00 & 0.00 & 0.00 & 0.00 & 0.00 & 0.00 & 0.00 & 0.00\\
\hline
\textsubarch{u}anjecaka & \textsubarch{u} & b & 0.06 & 0.23 & 0.28 & 0.16 & 0.06 & 0.14 & 0.02 & 0.02 & 0.02 & 0.01\\
\textsubarch{u}anjecaka & \textsubarch{u} & g & 1.00 & 0.00 & 0.00 & 0.00 & 0.00 & 0.00 & 0.00 & 0.00 & 0.00 & 0.00\\
\textsubarch{u}anjecaka & \textsubarch{u} & w & 1.00 & 0.00 & 0.00 & 0.00 & 0.00 & 0.00 & 0.00 & 0.00 & 0.00 & 0.00\\
\hline
\textsubarch{u}ar\={a}\'{\j}a & \textsubarch{u} & b & 0.03 & 0.21 & 0.41 & 0.15 & 0.09 & 0.04 & 0.02 & 0.02 & 0.02 & 0.01\\
\textsubarch{u}ar\={a}\'{\j}a & \textsubarch{u} & g & 1.00 & 0.00 & 0.00 & 0.00 & 0.00 & 0.00 & 0.00 & 0.00 & 0.00 & 0.00\\
\textsubarch{u}ar\={a}\'{\j}a & \textsubarch{u} & w & 1.00 & 0.00 & 0.00 & 0.00 & 0.00 & 0.00 & 0.00 & 0.00 & 0.00 & 0.00\\
\hline
\textsubarch{u}ara & \textsubarch{u} & b & 0.74 & 0.07 & 0.07 & 0.03 & 0.02 & 0.06 & 0.00 & 0.00 & 0.00 & 0.00\\
\textsubarch{u}ara & \textsubarch{u} & g & 1.00 & 0.00 & 0.00 & 0.00 & 0.00 & 0.00 & 0.00 & 0.00 & 0.00 & 0.00\\
\textsubarch{u}ara & \textsubarch{u} & w & 1.00 & 0.00 & 0.00 & 0.00 & 0.00 & 0.00 & 0.00 & 0.00 & 0.00 & 0.00\\
\hline
\textsubarch{u}arka & \textsubarch{u} & b & 0.72 & 0.07 & 0.09 & 0.04 & 0.03 & 0.03 & 0.01 & 0.00 & 0.00 & 0.00\\
\textsubarch{u}arka & \textsubarch{u} & w & 0.89 & 0.00 & 0.11 & 0.00 & 0.00 & 0.00 & 0.00 & 0.00 & 0.00 & 0.00\\
\textsubarch{u}arka & r & l & 0.00 & 0.25 & 0.36 & 0.19 & 0.08 & 0.04 & 0.03 & 0.02 & 0.02 & 0.01\\
\textsubarch{u}arka & r & r & 1.00 & 0.00 & 0.00 & 0.00 & 0.00 & 0.00 & 0.00 & 0.00 & 0.00 & 0.00\\
\hline
\textsubarch{u}arna-ka & \textsubarch{u} & b & 0.70 & 0.07 & 0.10 & 0.05 & 0.03 & 0.03 & 0.01 & 0.01 & 0.00 & 0.00\\
\textsubarch{u}arna-ka & \textsubarch{u} & g & 0.97 & 0.00 & 0.03 & 0.00 & 0.00 & 0.00 & 0.00 & 0.00 & 0.00 & 0.00\\
\textsubarch{u}arna-ka & \textsubarch{u} & w & 0.90 & 0.00 & 0.10 & 0.00 & 0.00 & 0.00 & 0.00 & 0.00 & 0.00 & 0.00\\
\textsubarch{u}arna-ka & rn & (r)r & 0.01 & 0.24 & 0.34 & 0.17 & 0.09 & 0.07 & 0.03 & 0.02 & 0.02 & 0.02\\
\textsubarch{u}arna-ka & rn & l & 1.00 & 0.00 & 0.00 & 0.00 & 0.00 & 0.00 & 0.00 & 0.00 & 0.00 & 0.00\\
\hline
\textsubarch{u}art- & \textsubarch{u} & g & 0.02 & 0.25 & 0.30 & 0.18 & 0.07 & 0.10 & 0.02 & 0.02 & 0.02 & 0.02\\
\textsubarch{u}art- & \textsubarch{u} & w & 1.00 & 0.00 & 0.00 & 0.00 & 0.00 & 0.00 & 0.00 & 0.00 & 0.00 & 0.00\\
\hline
\textsubarch{u}as\={i} & \textsubarch{u} & b & 0.81 & 0.05 & 0.05 & 0.02 & 0.01 & 0.04 & 0.00 & 0.00 & 0.00 & 0.00\\
\textsubarch{u}as\={i} & \textsubarch{u} & w & 0.62 & 0.11 & 0.12 & 0.08 & 0.03 & 0.01 & 0.01 & 0.01 & 0.01 & 0.00\\
\hline
\textsubarch{u}at-caka & \textsubarch{u} & b & 0.63 & 0.08 & 0.14 & 0.06 & 0.03 & 0.04 & 0.01 & 0.01 & 0.01 & 0.01\\
\textsubarch{u}at-caka & \textsubarch{u} & w & 1.00 & 0.00 & 0.00 & 0.00 & 0.00 & 0.00 & 0.00 & 0.00 & 0.00 & 0.00\\
\hline
\textsubarch{u}ata & \textsubarch{u} & b & 0.03 & 0.22 & 0.32 & 0.16 & 0.08 & 0.12 & 0.02 & 0.02 & 0.02 & 0.02\\
\textsubarch{u}ata & \textsubarch{u} & w & 1.00 & 0.00 & 0.00 & 0.00 & 0.00 & 0.00 & 0.00 & 0.00 & 0.00 & 0.00\\
\hline
\textsubarch{u}i\'{c}ati & \textsubarch{u} & b & 0.01 & 0.23 & 0.34 & 0.17 & 0.08 & 0.09 & 0.02 & 0.02 & 0.02 & 0.02\\
\textsubarch{u}i\'{c}ati & \textsubarch{u} & g & 1.00 & 0.00 & 0.00 & 0.00 & 0.00 & 0.00 & 0.00 & 0.00 & 0.00 & 0.00\\
\textsubarch{u}i\'{c}ati & \textsubarch{u} & w & 1.00 & 0.00 & 0.00 & 0.00 & 0.00 & 0.00 & 0.00 & 0.00 & 0.00 & 0.00\\
\hline
\textsubarch{u}i\textsubarch{i}ap\={a}na & \textsubarch{u} & b & 0.05 & 0.24 & 0.26 & 0.14 & 0.07 & 0.17 & 0.02 & 0.02 & 0.02 & 0.01\\
\textsubarch{u}i\textsubarch{i}ap\={a}na & \textsubarch{u} & g & 1.00 & 0.00 & 0.00 & 0.00 & 0.00 & 0.00 & 0.00 & 0.00 & 0.00 & 0.00\\
\textsubarch{u}i\textsubarch{i}ap\={a}na & \textsubarch{u} & w & 1.00 & 0.00 & 0.00 & 0.00 & 0.00 & 0.00 & 0.00 & 0.00 & 0.00 & 0.00\\
\hline
\textsubarch{u}ic\={a}ra & \textsubarch{u} & g & 0.69 & 0.08 & 0.09 & 0.05 & 0.03 & 0.05 & 0.01 & 0.01 & 0.01 & 0.00\\
\hline
\textsubarch{u}id\={a}na & \textsubarch{u} & b & 0.57 & 0.10 & 0.10 & 0.08 & 0.03 & 0.08 & 0.01 & 0.01 & 0.01 & 0.01\\
\textsubarch{u}id\={a}na & \textsubarch{u} & g & 0.98 & 0.00 & 0.02 & 0.00 & 0.00 & 0.00 & 0.00 & 0.00 & 0.00 & 0.00\\
\textsubarch{u}id\={a}na & \textsubarch{u} & w & 0.90 & 0.00 & 0.10 & 0.00 & 0.00 & 0.00 & 0.00 & 0.00 & 0.00 & 0.00\\
\hline
\textsubarch{u}ida\textsubarch{u}\={a} & \textsubarch{u} & b & 0.02 & 0.24 & 0.33 & 0.17 & 0.09 & 0.08 & 0.02 & 0.02 & 0.02 & 0.01\\
\textsubarch{u}ida\textsubarch{u}\={a} & \textsubarch{u} & w & 1.00 & 0.00 & 0.00 & 0.00 & 0.00 & 0.00 & 0.00 & 0.00 & 0.00 & 0.00\\
\hline
\textsubarch{u}in\={a}\'{c}a & \'{c} & h & 0.74 & 0.06 & 0.09 & 0.04 & 0.02 & 0.03 & 0.01 & 0.00 & 0.00 & 0.00\\
\textsubarch{u}in\={a}\'{c}a & \textsubarch{u} & g & 0.02 & 0.22 & 0.34 & 0.15 & 0.08 & 0.12 & 0.02 & 0.02 & 0.02 & 0.01\\
\textsubarch{u}in\={a}\'{c}a & \textsubarch{u} & w & 1.00 & 0.00 & 0.00 & 0.00 & 0.00 & 0.00 & 0.00 & 0.00 & 0.00 & 0.00\\
\hline
\textsubarch{u}rinji & \textsubarch{u} & b & 0.72 & 0.07 & 0.09 & 0.05 & 0.02 & 0.02 & 0.01 & 0.00 & 0.00 & 0.00\\
\textsubarch{u}rinji & \textsubarch{u} & g & 0.81 & 0.05 & 0.05 & 0.02 & 0.01 & 0.04 & 0.00 & 0.00 & 0.00 & 0.00\\
\hline
a\'{c}\textsubarch{u}a & \'{c}\textsubarch{u} & s & 0.54 & 0.11 & 0.14 & 0.09 & 0.02 & 0.07 & 0.01 & 0.01 & 0.01 & 0.01\\
a\'{c}\textsubarch{u}a & \'{c}\textsubarch{u} & sp & 0.71 & 0.07 & 0.10 & 0.05 & 0.02 & 0.02 & 0.01 & 0.01 & 0.00 & 0.00\\
\hline
a\'{c}ru & \'{c}r & rs & 0.02 & 0.24 & 0.28 & 0.17 & 0.08 & 0.14 & 0.02 & 0.02 & 0.02 & 0.02\\
a\'{c}ru & \'{c}r & sr & 1.00 & 0.00 & 0.00 & 0.00 & 0.00 & 0.00 & 0.00 & 0.00 & 0.00 & 0.00\\
a\'{c}ru & pro & no pro & 0.26 & 0.18 & 0.21 & 0.12 & 0.06 & 0.12 & 0.02 & 0.01 & 0.01 & 0.01\\
a\'{c}ru & pro & pro & 1.00 & 0.00 & 0.00 & 0.00 & 0.00 & 0.00 & 0.00 & 0.00 & 0.00 & 0.00\\
\hline
ar\v{s}a & pro & no pro & 0.00 & 0.18 & 0.50 & 0.12 & 0.10 & 0.06 & 0.01 & 0.01 & 0.01 & 0.01\\
ar\v{s}a & pro & pro & 1.00 & 0.00 & 0.00 & 0.00 & 0.00 & 0.00 & 0.00 & 0.00 & 0.00 & 0.00\\
ar\v{s}a & rs & \v{s} & 0.67 & 0.10 & 0.06 & 0.08 & 0.05 & 0.01 & 0.01 & 0.01 & 0.01 & 0.01\\
ar\v{s}a & rs & rch & 0.43 & 0.00 & 0.57 & 0.00 & 0.00 & 0.00 & 0.00 & 0.00 & 0.00 & 0.00\\
ar\v{s}a & rs & rs & 0.63 & 0.00 & 0.37 & 0.00 & 0.00 & 0.00 & 0.00 & 0.00 & 0.00 & 0.00\\
\hline
b\textsubring{r}\'{\j}ant & r\'{\j} & l & 0.58 & 0.11 & 0.15 & 0.08 & 0.04 & 0.02 & 0.01 & 0.01 & 0.01 & 0.01\\
b\textsubring{r}\'{\j}ant & r\'{\j} & r\'{\j} & 1.00 & 0.00 & 0.00 & 0.00 & 0.00 & 0.00 & 0.00 & 0.00 & 0.00 & 0.00\\
\hline
barda & rd & l & 0.02 & 0.23 & 0.35 & 0.16 & 0.08 & 0.10 & 0.02 & 0.02 & 0.01 & 0.01\\
barda & rd & rd & 1.00 & 0.00 & 0.00 & 0.00 & 0.00 & 0.00 & 0.00 & 0.00 & 0.00 & 0.00\\
\hline
bar\'{\j}\={a}d & r\'{\j} & l & 0.68 & 0.08 & 0.12 & 0.05 & 0.03 & 0.02 & 0.01 & 0.01 & 0.00 & 0.00\\
bar\'{\j}\={a}d & r\'{\j} & r\'{\j} & 1.00 & 0.00 & 0.00 & 0.00 & 0.00 & 0.00 & 0.00 & 0.00 & 0.00 & 0.00\\
\hline
ca{\texttheta}\textsubarch{u}\textsubring{r}\'{c}at & r\'{c} & l & 0.00 & 0.25 & 0.34 & 0.18 & 0.08 & 0.07 & 0.03 & 0.02 & 0.02 & 0.02\\
ca{\texttheta}\textsubarch{u}\textsubring{r}\'{c}at & r\'{c} & r\'{c} & 1.00 & 0.00 & 0.00 & 0.00 & 0.00 & 0.00 & 0.00 & 0.00 & 0.00 & 0.00\\
\hline
da\'{c}a & \'{c} & h & 0.71 & 0.07 & 0.10 & 0.05 & 0.02 & 0.02 & 0.01 & 0.01 & 0.00 & 0.00\\
da\'{c}a & \'{c} & s & 1.00 & 0.00 & 0.00 & 0.00 & 0.00 & 0.00 & 0.00 & 0.00 & 0.00 & 0.00\\
\hline
darn & rn & (r)n & 0.03 & 0.20 & 0.41 & 0.11 & 0.09 & 0.12 & 0.01 & 0.01 & 0.01 & 0.01\\
darn & rn & (r)r & 1.00 & 0.00 & 0.00 & 0.00 & 0.00 & 0.00 & 0.00 & 0.00 & 0.00 & 0.00\\
\hline
g(a)rna-ka & rn & (r)n & 0.02 & 0.15 & 0.53 & 0.10 & 0.11 & 0.06 & 0.01 & 0.01 & 0.01 & 0.01\\
g(a)rna-ka & rn & l & 1.00 & 0.00 & 0.00 & 0.00 & 0.00 & 0.00 & 0.00 & 0.00 & 0.00 & 0.00\\
\hline
gau-\'{c}\textsubarch{u}anta- & \'{c}\textsubarch{u} & s & 0.82 & 0.05 & 0.05 & 0.04 & 0.01 & 0.00 & 0.00 & 0.00 & 0.00 & 0.00\\
gau-\'{c}\textsubarch{u}anta- & \'{c}\textsubarch{u} & sp & 0.76 & 0.06 & 0.08 & 0.04 & 0.02 & 0.02 & 0.01 & 0.00 & 0.00 & 0.00\\
\hline
hi\'{\j}\textsubarch{u}\={a}na & \'{\j}\textsubarch{u} & zb & 0.01 & 0.25 & 0.32 & 0.18 & 0.06 & 0.10 & 0.03 & 0.02 & 0.02 & 0.02\\
hi\'{\j}\textsubarch{u}\={a}na & \'{\j}\textsubarch{u} & zm & 1.00 & 0.00 & 0.00 & 0.00 & 0.00 & 0.00 & 0.00 & 0.00 & 0.00 & 0.00\\
hi\'{\j}\textsubarch{u}\={a}na & \'{\j}\textsubarch{u} & zw & 1.00 & 0.00 & 0.00 & 0.00 & 0.00 & 0.00 & 0.00 & 0.00 & 0.00 & 0.00\\
\hline
k\textsubring{r}mi & r & l & 0.58 & 0.11 & 0.11 & 0.08 & 0.02 & 0.06 & 0.01 & 0.01 & 0.01 & 0.01\\
k\textsubring{r}mi & r & r & 0.69 & 0.08 & 0.11 & 0.06 & 0.02 & 0.02 & 0.01 & 0.01 & 0.00 & 0.00\\
\hline
ka\'{c}\textsubarch{i}apa & \'{c}\textsubarch{i} & \v{s} & 0.03 & 0.24 & 0.28 & 0.15 & 0.06 & 0.18 & 0.02 & 0.02 & 0.01 & 0.01\\
ka\'{c}\textsubarch{i}apa & \'{c}\textsubarch{i} & s & 1.00 & 0.00 & 0.00 & 0.00 & 0.00 & 0.00 & 0.00 & 0.00 & 0.00 & 0.00\\
\hline
laup\={a}\'{c}a & \'{c} & h & 0.64 & 0.09 & 0.14 & 0.06 & 0.03 & 0.02 & 0.01 & 0.01 & 0.01 & 0.01\\
laup\={a}\'{c}a & \'{c} & s & 1.00 & 0.00 & 0.00 & 0.00 & 0.00 & 0.00 & 0.00 & 0.00 & 0.00 & 0.00\\
laup\={a}\'{c}a & l & l & 0.59 & 0.11 & 0.14 & 0.07 & 0.04 & 0.02 & 0.01 & 0.01 & 0.01 & 0.01\\
laup\={a}\'{c}a & l & r & 1.00 & 0.00 & 0.00 & 0.00 & 0.00 & 0.00 & 0.00 & 0.00 & 0.00 & 0.00\\
\hline
mu\v{s}ti & \v{s}t & \v{s}t & 0.53 & 0.12 & 0.17 & 0.08 & 0.04 & 0.03 & 0.01 & 0.01 & 0.01 & 0.01\\
mu\v{s}ti & \v{s}t & st & 1.00 & 0.00 & 0.00 & 0.00 & 0.00 & 0.00 & 0.00 & 0.00 & 0.00 & 0.00\\
\hline
p\textsubring{r}tu & rt & l & 0.58 & 0.13 & 0.11 & 0.09 & 0.04 & 0.03 & 0.01 & 0.01 & 0.01 & 0.01\\
p\textsubring{r}tu & rt & rt & 0.91 & 0.00 & 0.09 & 0.00 & 0.00 & 0.00 & 0.00 & 0.00 & 0.00 & 0.00\\
\hline
parna & rn & (r)n & 0.01 & 0.22 & 0.29 & 0.16 & 0.12 & 0.13 & 0.02 & 0.02 & 0.02 & 0.02\\
parna & rn & (r)r & 1.00 & 0.00 & 0.00 & 0.00 & 0.00 & 0.00 & 0.00 & 0.00 & 0.00 & 0.00\\
parna & rn & l & 1.00 & 0.00 & 0.00 & 0.00 & 0.00 & 0.00 & 0.00 & 0.00 & 0.00 & 0.00\\
\hline
pu{\texttheta}ra & tr & s & 0.70 & 0.08 & 0.09 & 0.05 & 0.02 & 0.04 & 0.01 & 0.01 & 0.01 & 0.00\\
pu{\texttheta}ra & tr & tr & 1.00 & 0.00 & 0.00 & 0.00 & 0.00 & 0.00 & 0.00 & 0.00 & 0.00 & 0.00\\
\hline
r\={a}\v{s}ta & \v{s}t & \v{s}t & 0.00 & 0.25 & 0.38 & 0.18 & 0.10 & 0.02 & 0.03 & 0.02 & 0.02 & 0.01\\
r\={a}\v{s}ta & \v{s}t & st & 1.00 & 0.00 & 0.00 & 0.00 & 0.00 & 0.00 & 0.00 & 0.00 & 0.00 & 0.00\\
\hline
skar & r & l & 0.09 & 0.20 & 0.24 & 0.16 & 0.12 & 0.11 & 0.02 & 0.02 & 0.02 & 0.02\\
skar & r & r & 1.00 & 0.00 & 0.00 & 0.00 & 0.00 & 0.00 & 0.00 & 0.00 & 0.00 & 0.00\\
\hline
sp\textsubring{r}\'{\j}an & r\'{\j} & l & 0.37 & 0.14 & 0.22 & 0.12 & 0.05 & 0.05 & 0.01 & 0.01 & 0.01 & 0.01\\
sp\textsubring{r}\'{\j}an & r\'{\j} & r\'{\j} & 1.00 & 0.00 & 0.00 & 0.00 & 0.00 & 0.00 & 0.00 & 0.00 & 0.00 & 0.00\\
\hline
x\v{s}\textsubarch{u}ifta- & t & r & 0.62 & 0.11 & 0.13 & 0.08 & 0.03 & 0.01 & 0.01 & 0.01 & 0.01 & 0.00\\
x\v{s}\textsubarch{u}ifta- & t & t & 0.57 & 0.13 & 0.14 & 0.08 & 0.04 & 0.01 & 0.01 & 0.01 & 0.01 & 0.01\\

\end{longtable}
}

\subsection*{Posterior distributions over dialect components for sound change instances, raw values}
\url{https://github.com/chundrac/w_ir_layers/blob/master/p_z_all} gives $P(z_i|\boldsymbol \theta,\boldsymbol \phi)$ for every sound change instance in our data set.


\end{document}